\documentclass[12pt,a4paper]{article}

\usepackage{fullpage}

\usepackage{cite}

\usepackage{graphicx}

\usepackage{amsmath}
\usepackage{array}

\usepackage{chngpage}

\usepackage{subfig}

\usepackage{overpic}
\usepackage[T1]{fontenc}
\usepackage[utf8x]{inputenc}
\usepackage[english]{babel}
\usepackage{color}

	\newlength{\figurewidth}
	\newlength{\figurewidthA}
	\newlength{\figurewidthB}
	\newlength{\figurewidthNY}
\begin{document}

\title{Using \emph{Robust PCA} to estimate regional characteristics of language use from geo-tagged Twitter messages}

\author{
D\'aniel Kondor, Istv\'an Csabai, L\'aszl\'o Dobos, J\'anos Sz\"ule, Norbert Barankai, \\
Tam\'as Hanyecz, Tam\'as Seb\H{o}k, Zs\'ofia Kallus and G\'abor Vattay \\
Department of Physics of Complex Systems, E\"otv\"os Lor\'and University\\
H-1117, P\'azm\'any P\'eter s\'et\'any 1/A, Budapest, Hungary\\
Email: kdani88@elte.hu
}

\maketitle

\begin{abstract}
	Principal component analysis (PCA) and related techniques have been successfully employed in natural language processing.
	Text mining applications in the age of the online social media (OSM) face new challenges due to properties specific to
	these use cases (e.g.~spelling issues specific to texts posted by users, the presence of spammers and bots, service
	announcements, etc.). In this paper, we employ a Robust PCA technique to separate typical outliers and highly
	localized topics from the low-dimensional structure present in language use in online social networks.
	Our focus is on identifying geospatial features among the messages posted by the users of the Twitter microblogging
	service. Using a dataset which consists of over 200 million geolocated tweets collected over the course of a year,
	we investigate whether the information present in word usage frequencies can be used to identify regional features of
	language use and topics of interest. Using the \emph{PCA pursuit} method, we are able to identify important
	low-dimensional features, which constitute smoothly varying functions of the geographic location.
\end{abstract}

\section{Introduction}

With the rapidly growing corpus of digitally available textual data, there has been a significant interest in unsupervised language
processing techniques~\cite{semantic}. Principal component analysis (PCA) and the closely related latent semantic
analysis (LSA)~\cite{lsalandauer} have been widely and successfully applied to various classes of documents to identify relations
among documents, e.g.~cluster the documents according to topics or indexing documents~\cite{lsaindexing,lsa1,lsa2}. Nowdays, LSA is employed by
many search engines to improve results based on clustering documents to topics of interest. While the basic recipe for LSA is
well known, each data source might require special treatment according to its unique characteristic.

A relatively new and potentially valuable source of digital textual data is the corpus of messages posted by users to online social network (OSN)
websites. Using these new datasources, previous research showed the possibility to gain valuable new
insights into the structure and dynamics of society, gaining the possibility to address questions whose study was previously limited by the
scarcity of available data~\cite{findmefacebook,localization1,diffusion1,diffusion2,postagging,memetracking}.
Also, the possibility to simultaneously analyze both the network topology of connections between users and the corpus of textual data
posted by them opens up new possibilities in the fields of social, network and computer sciences~\cite{leskovec,watts,influence1,influence2}.
A valuable aspect of some online social networks is the inclusion of spatial data. For example, the users of the Twitter microblogging service
have the ability to attach their geographic coordinates to their short messages (tweets), which opens up the possibility to analyze
the geographical variation of language use on massive amounts of readily available data~\cite{diffusion2,localization1}.

In this paper we evaluate principal component analysis (PCA) techniques on a dataset which consists of geo-tagged Twitter messages (tweets),
and investigate the feasibility of identifying regional characteristics of language use.
Our approach is to consider a medium-sized geographical region as one ``document'',
and apply PCA to the corpus of documents which we produce by assigning the text of tweets to the to the geographic region
where it was posted. In this way, the rows of our term-document matrix correspond to geographical regions and the columns to the words
included in our corpus. This method ignores the possible variation
present inside the regions (e.g.~multiple distinct topics in one region), and the possibility to detect topics which are globally present. This
is in accordance in our goals; we wish to identify the regional variations. Detection of topics without a clear regional focus
might be possible with other natural language processing methods, and local
variations can be detected by ``zooming in'' to regions with a high number of tweets to obtain a finer-grained picture of language use.

In many real-world PCA applications, the data matrix can be considered as the weighted sum of two parts, one of which is sparse
and the other is low-rank. The low-rank part can be thought of as some
``background'' component, while the sparse part can either contain the relevant data or can be composed of some nontrivial outliers.
Depending on the nature of the problem studied, either one or both of them might be of interest. In these
cases, it would be desirable to be able to separate these two parts, and to analyze them separately. Also, if the sparse part has a large
magnitude, the principal components found by PCA will be dominated by the outliers and revealing the low-rank
structure present in the data can prove difficult. In the case of our corpus of Twitter messages, we expect the data matrix to be indeed
separable into these two parts. We expect the low-rank component to represent the true variation in language use (e.g.~usage frequencies
of words will possibly be different in different geographic regions), and the sparse part to contain highly localized topics of interest
(e.g.~landmarks). To deal with these issues, we employ the \emph{Principal Component Pursuit} technique developed by Cand\`es~et.al.,
which achieves the separation of the original data matrix into its sparse and low-rank parts effectively~\cite{pcapursuit1,pcapursuit2}.

\section{Methods}

\subsection{Twitter data}

We used the publicly available sample datastream to download geotagged tweets. Our dataset includes a total of 1.3 billion
tweets between August 2012 and March 2013. Among these, 725 million are geo-tagged, i.e.~have GPS coordinated
associated with them. In this paper, we limit our analysis to tweets from the contiguous United Stated of America, a
total of approximately 245 million tweets.
We constructed a geographically indexed database of these tweets, which allows the efficient analysis of regional
features~\cite{twitterdb}. We use the HTM scheme for geographic indexing~\cite{htm}. This employs a quad-tree, where
the surface of the Earth is recursively divided into spherical triangles called HTM cells or \emph{trixels}. The
indexing works on multiple levels; a deeper (i.e.~more detailed) level can be acquired by dividing each trixel into
4 smaller triangles by connecting the midpoints of the sides. The subdivision is started by 8 spherical triangles (level 0).

\begin{figure}
	\centering
	\begin{overpic}[width=\figurewidthA
		]{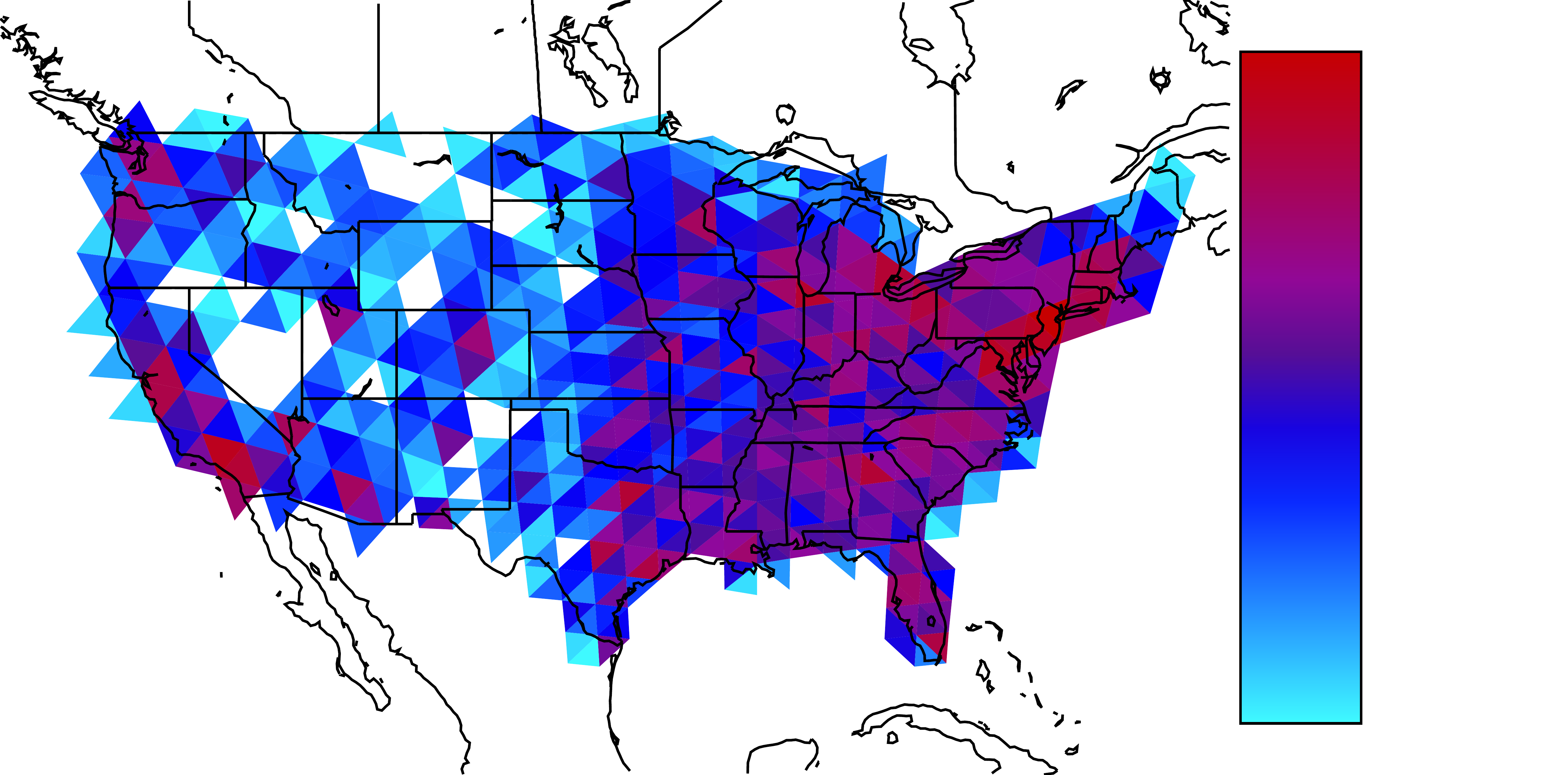}
		\put(89,45){$\scriptstyle 50000000$}
		\put(89,36.4){$\scriptstyle 10000000$}
		\put(89,27.8){$\scriptstyle 1700000$}
		\put(89,19.2){$\scriptstyle 300000$}
		\put(89,10.6){$\scriptstyle 55000$}
		\put(89,2) {$\scriptstyle 10000$}
	\end{overpic}
	\caption{Number of words from the corpus in the trixels included in the analysis. Cells with less than 10000 occurrences of at least 1000
		individual words were excluded from our analysis; these are left blank here too. Note that the colorscale is logarithmic.}
	\label{scntl}
\end{figure}

We compile the list of the most frequently used words from the tweets, and compute their spatial distribution. We divide the
USA to level 6 HTM cells; we have a total of 558 cells with an approximate cell area of 15500 km$^2$ each. We construct
the word occurrence matrix $W_{ij}$ as the number of occurrences of the $i$-th word in the $j$-th cell. As the population density
of the USA is very heterogeneous, the number of word occurrences in each cell is also heterogeneous. To improve the quality of
the dataset, we only include cells which contain at least 10000 occurrences of at least 1000 individual words. We also
limit the words used to those with at least 10000 occurrences in at least 300 individual cells. This way there remain
491 cells (see Fig.~\ref{scntl}) and 6032 words, which form the $W_{ij}$ word occurrence matrix. Before
applying the PCA procedure, we normalize $W_{ij}$ so that the elements are the relative frequencies of words:
$X_{ij} \equiv W_{ij} / \sum_k W_{kj}$, i.e.~we normalize each element by the total number of words posted in that cell.
Due to the structure of our data matrix, and the employed Robust PCA method, we chose not to subtract averages
from the data; of course, this will probably result in that average word frequencies dominate the first principal
components.

\begin{figure}
	\centering
	\begin{overpic}[width=10cm
		]{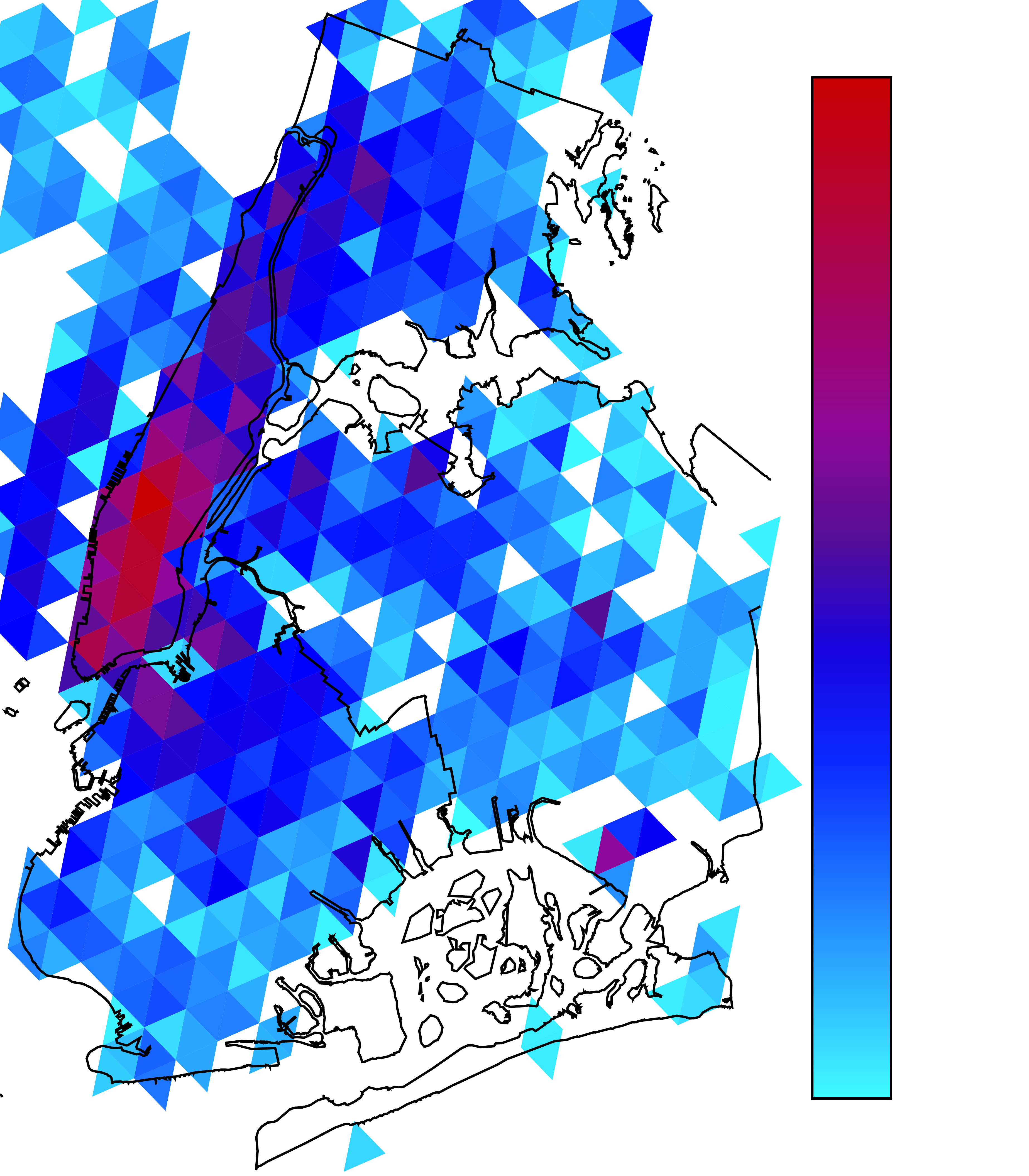}
		\put(77,92.5){$\scriptstyle 1200000$}
		\put(77,75.2){$\scriptstyle 425000$}
		\put(77,57.9){$\scriptstyle 150000$}
		\put(77,40.6){$\scriptstyle 52000$}
		\put(77,23.3){$\scriptstyle 18000$}
		\put(77,6){$\scriptstyle 6500$}
	\end{overpic}
	\caption{Number of words from the corpus in the trixels included for New York City. Cells with less than 6500 occurrences of at least 1000
		individual words were excluded. Note that the colorscale is logarithmic.}
	\label{nycntl}
\end{figure}

Furthermore, we compile a higher resolution data matrix from the same dataset focusing on New York City, with a total of
6.3 million tweets. In this case, we use level
13 HTM cells (approximate cell area: $0.95\,\textrm{km}^2$), to possibly identify small-scale language use differences.
We limit our analysis to trixels with at least $6500$ occurrences of at least $1000$ individual words. Also, we only include words
with at least $1000$ occurrences. This way, we have $525$ trixels and $3979$ words (see Fig.~\ref{nycntl}).

\subsection{Classic PCA approach}

We compute the singular value decomposition (SVD) of the data matrix $X$:
\begin{equation}
	X = U \Sigma V^{T}
\end{equation}
The diagonal matrix $\Sigma$ contains the singular values, while the column of $U$ and $V$ contain the principal components.
The components reveal the sources of variation in the data, ordered by their magnitude of contribution.

To analyze the results we can plot each component (i.e.~the columns of the $V$ matrix) on the map of the USA; we color each
trixel according to the corresponding value (see Fig.~\ref{twV5}). We also inspect the values which can be associated to
the individual words in each component (i.e.~the columns of the $U$ matrix).

\subsection{Robust PCA}

While the classic PCA method is useful for detecting structure in the data, in some cases, it might not give optimal results.
One such case is that when there are outliers in the data; in this case, the main principal components are dominated by them
and identifying a possible low-rank structure becomes more challenging. Filtering out these can also prove difficult, as in
many cases, it is not straightforward to estimate the relevance of components in the raw data. In the
case of Twitter messages, this would mean the identification of the sources of outliers (e.g.~spammers and advertisement,
weather forecast, local tourist attractions), and excluding messages containing them. While this would improve the results
of the PCA, it might also leave out some relevant data. A more favorable approach would be a method, which can be applied to
the raw data matrix, and automatically separates outliers (i.e.~sparse parts) and low-dimensional structure.
In this case, we can inspect the sparse part separately, and decide which components are relevant.
This can be obtained by the \emph{Principal Component Pursuit} technique~\cite{pcapursuit1,pcapursuit2}, which separates the data
matrix into these two parts:
\begin{equation}
	X = X^{\textrm{S}} + X^{\textrm{LR}} \, \textrm{,}
\end{equation}
where $X^{\textrm{S}}$ is a sparse matrix and $X^{\textrm{LR}}$ contains the dense but low-rank part of the data.
This is achieved by minimizing the sum
\begin{equation}
	\lambda \| X^{\textrm{S}} \|_1 + \| X^{\textrm{LR}} \|_{\sigma} \, \textrm{,}
\end{equation}
where for a matrix $X$ of dimensions $n_1 \times n_2$ with $n_1 \geq n_2$, $\lambda = 1 / \sqrt{n_1}$, and the norms are the
$l_1$ and nuclear norms respectively:
\begin{gather}
\begin{split}
	\| X \|_1 = \sum_{ij} |X_{ij}| \\
	\| X \|_{\sigma} = \sum_i \sigma_i (X) \, \textrm{.}
\end{split}
\end{gather}
Here $\sigma_i (X)$ denotes the $i$-th singular value of $X$. To obtain the results we use the Matlab code developed by
Lin~et~al. implementing the inexact augmented Lagrangian method~\cite{pcapursuit1,alm}.

We carry out the SVD for both matrices and analyze the results simultaneously.
We expect the principal components of $X^{\textrm{S}}$ to contain information about highly localized trends (i.e.~features
specific to only one or few cells), and the components of $X^{\textrm{LR}}$ to contain the possibly smooth variations in language
use.

\section{Results}

\subsection{USA}

\begin{figure}
	\centering
		\begin{overpic}[width=\figurewidthB
			]{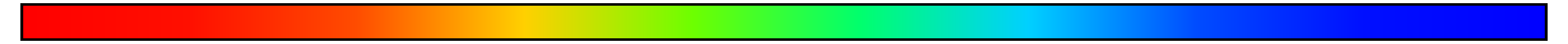}
			\put(0.5,-1){$-$}
			\put(98,-1){$+$}
		\end{overpic}
	\\
	\subfloat[Classic PCA, first component]{
		\includegraphics[width=\figurewidth ]{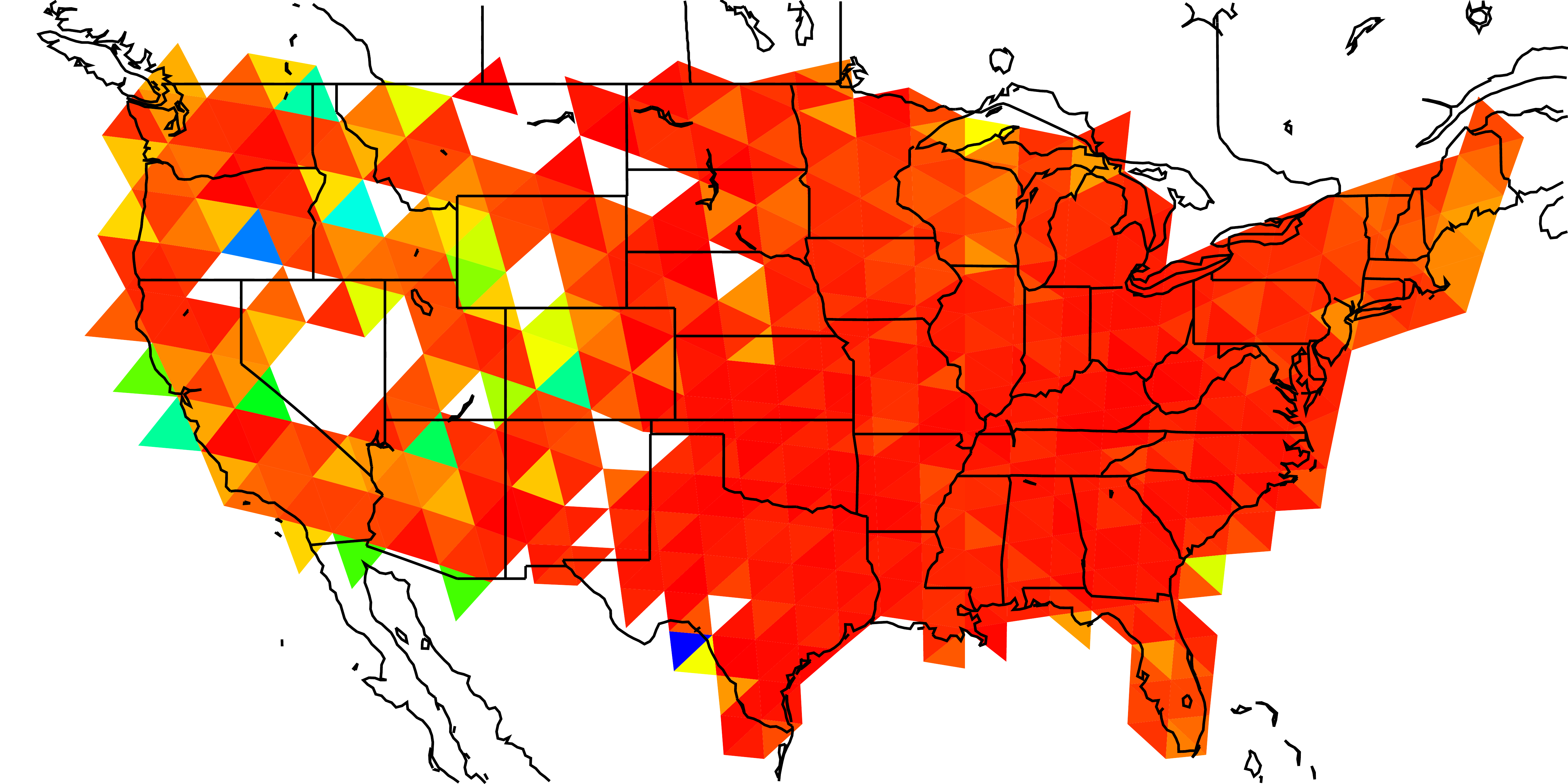}
	}
	\quad
	\subfloat[PCA Pursuit, sparse part, first component]{
		\includegraphics[width=\figurewidth ]{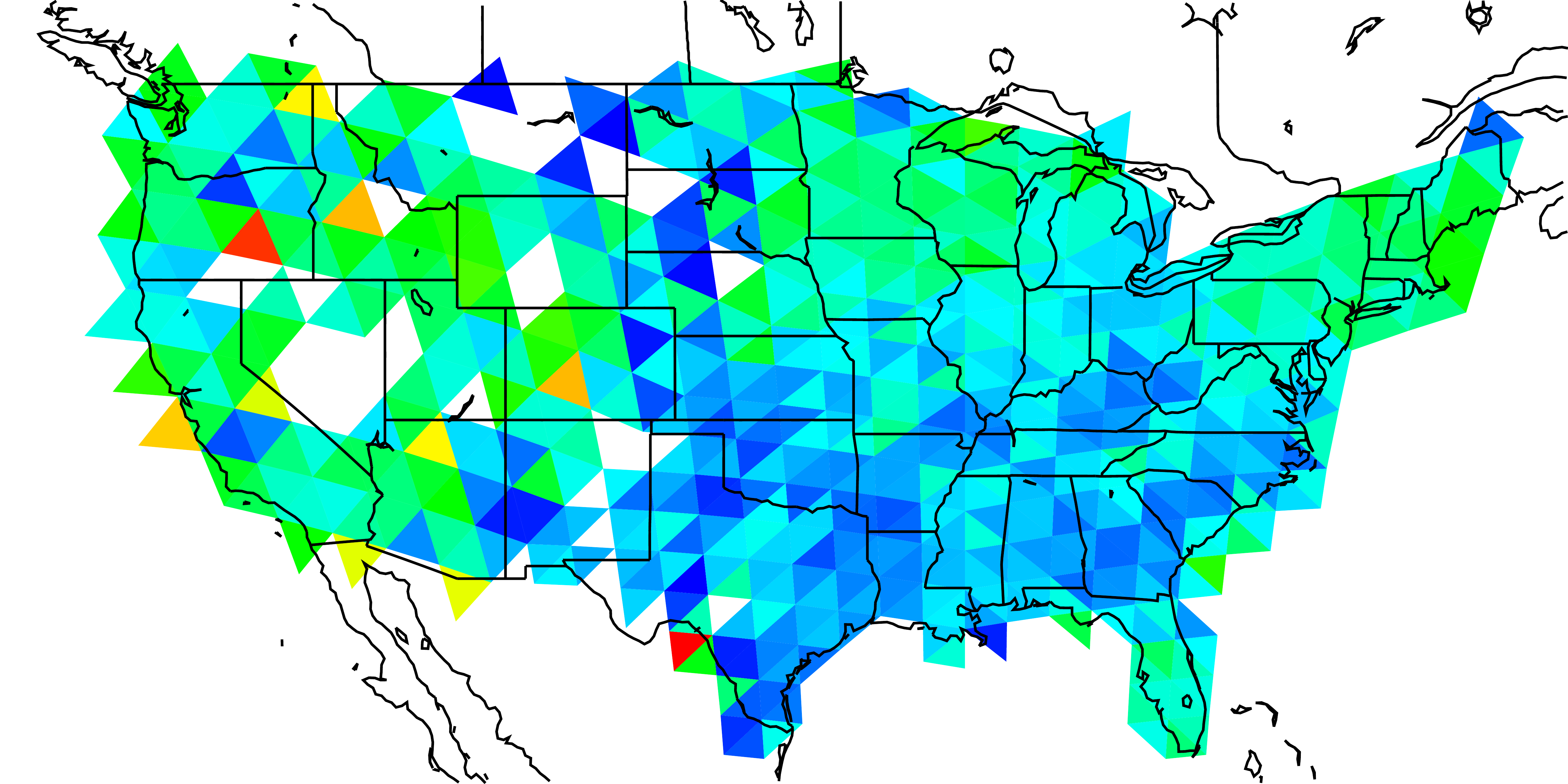}
	}
	\quad
	\subfloat[PCA Pursuit, low rank part, first component]{
		\includegraphics[width=\figurewidth ]{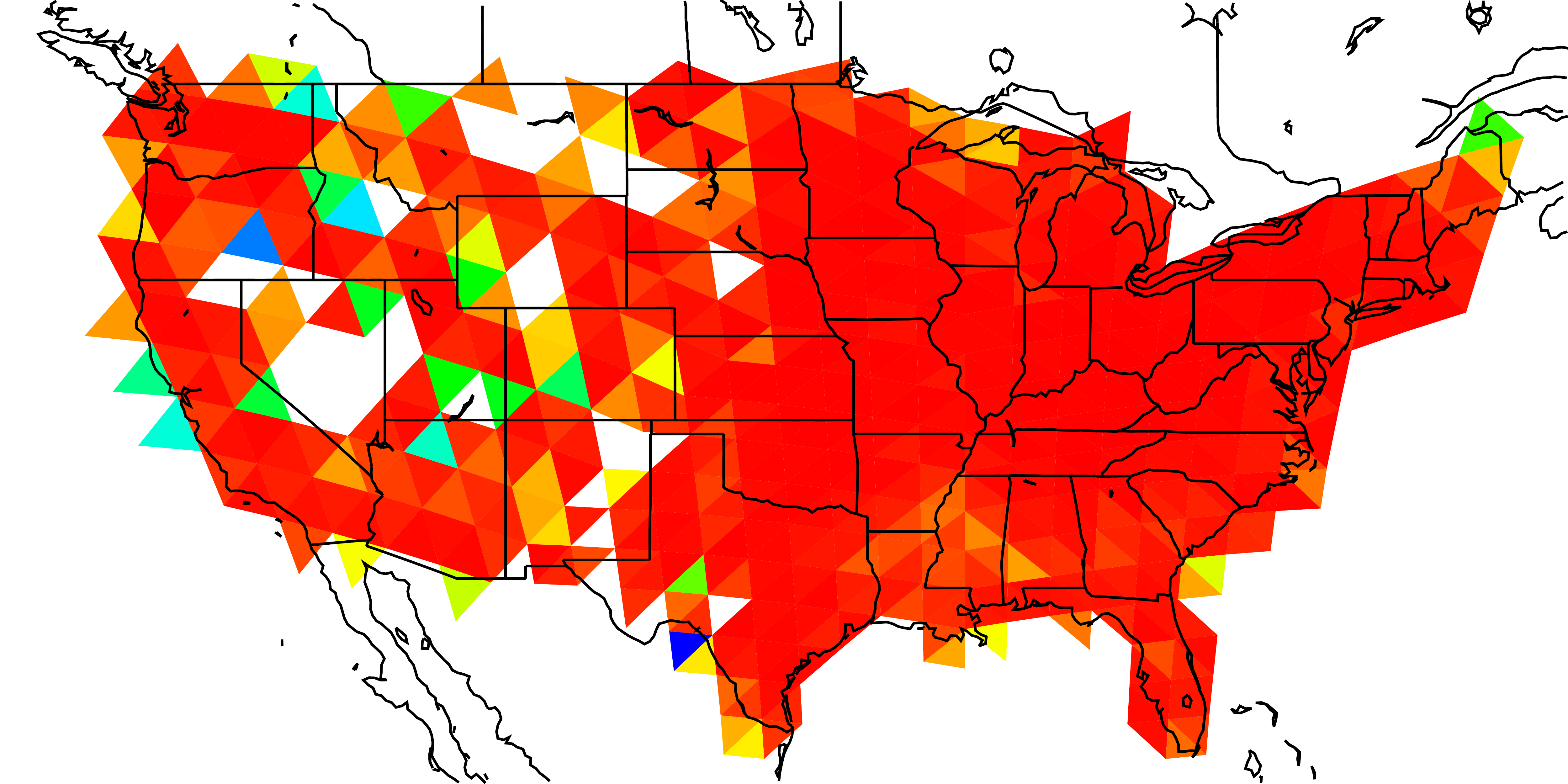}
	}
	\\
		\subfloat[Classic PCA, second component]{
		\includegraphics[width=\figurewidth ]{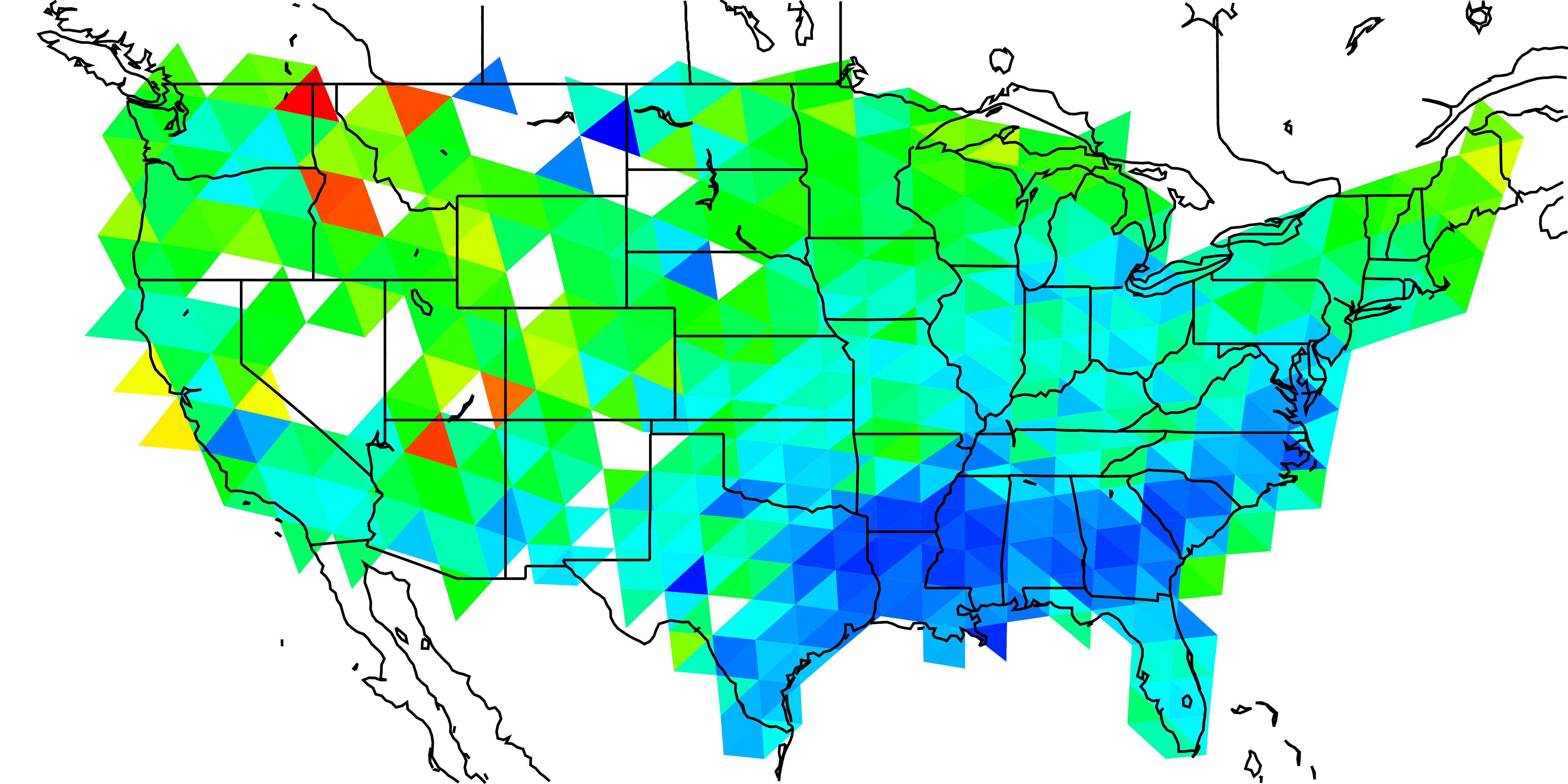}
	}
	\quad
	\subfloat[PCA Pursuit, sparse part, second component]{
		\includegraphics[width=\figurewidth ]{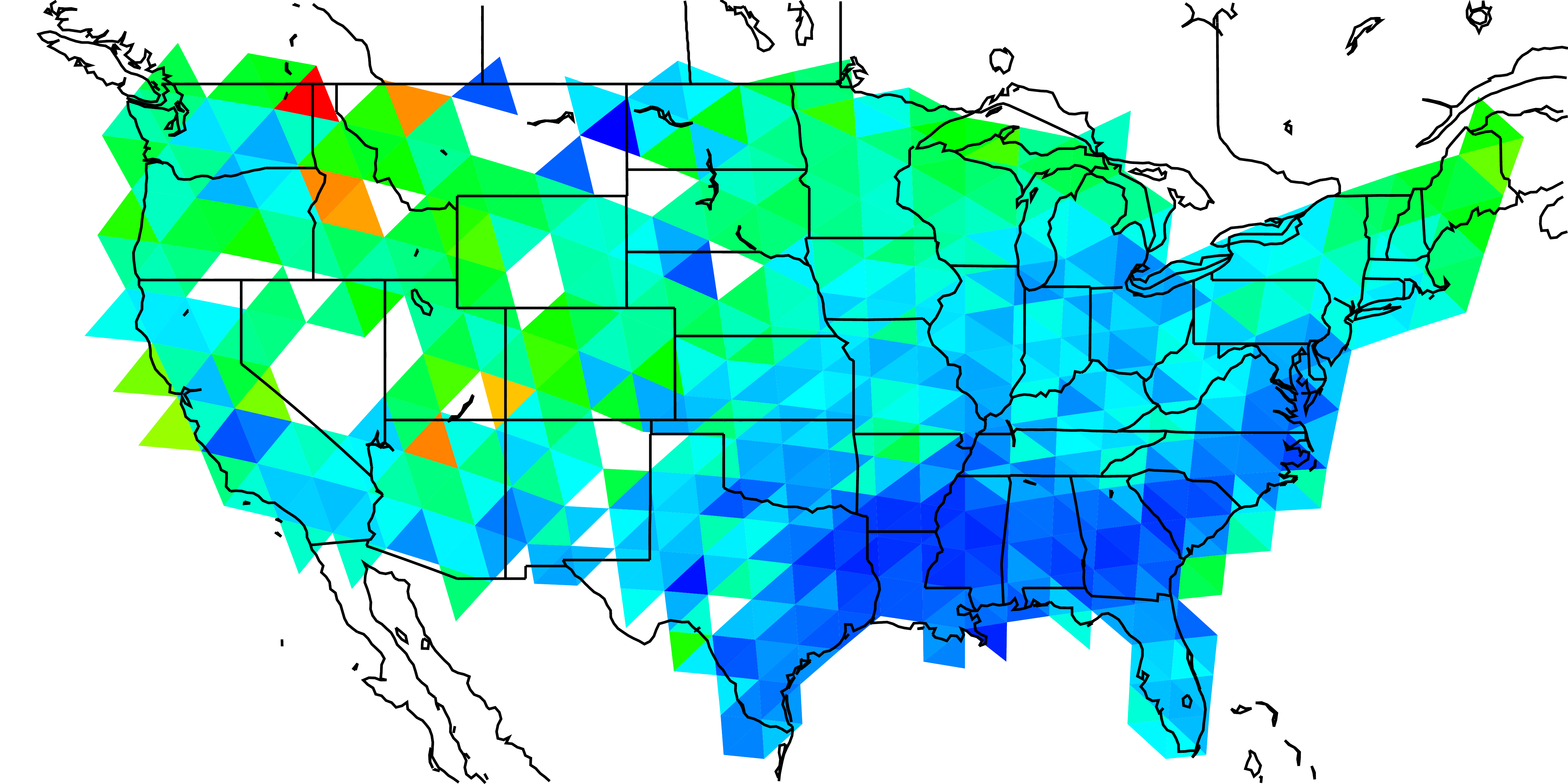}
	}
	\quad
	\subfloat[PCA Pursuit, low rank part, second component]{
		\includegraphics[width=\figurewidth ]{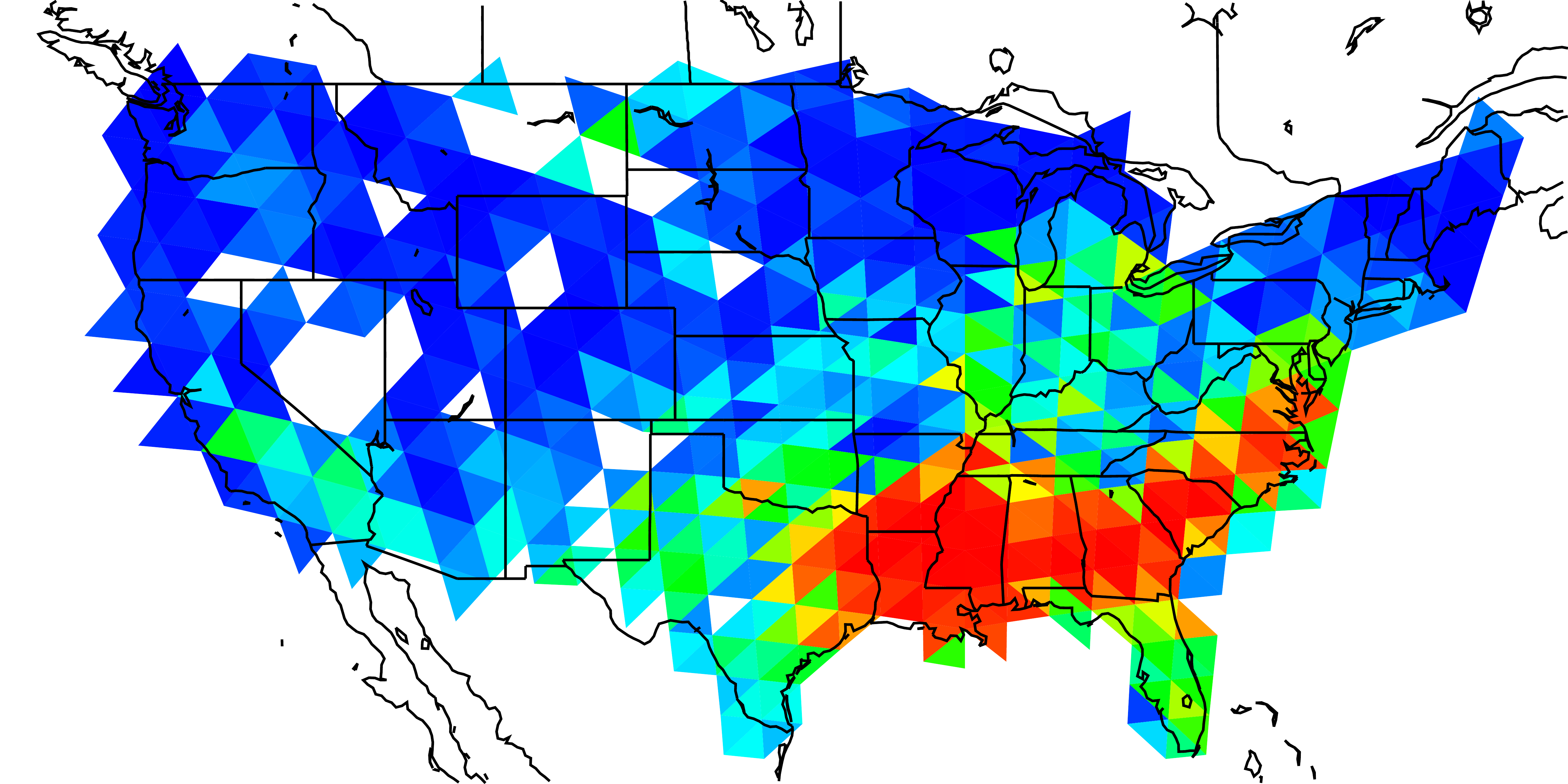}
	}
	\\
		\subfloat[Classic PCA, third component]{
		\includegraphics[width=\figurewidth ]{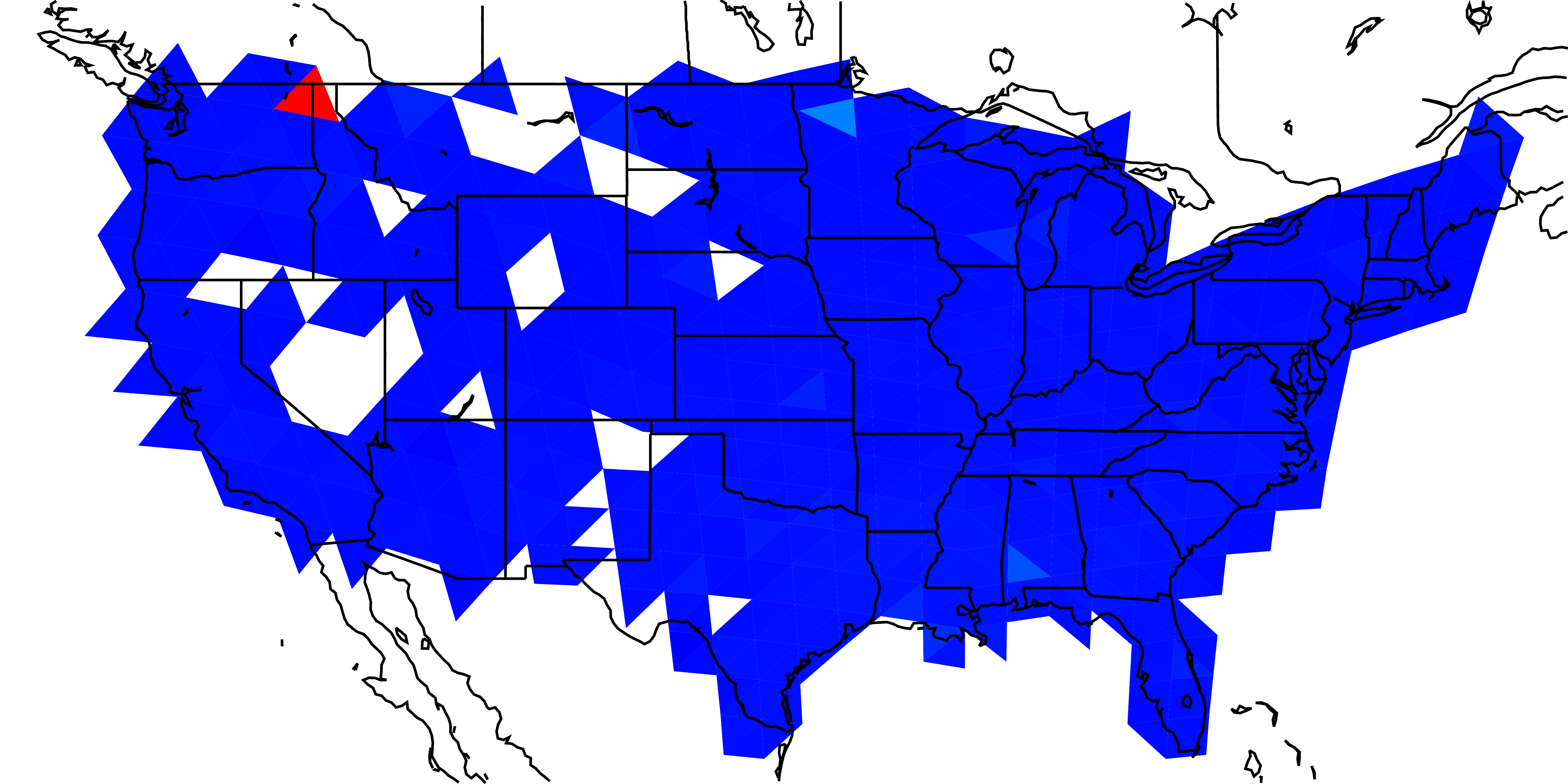}
	}
	\quad
	\subfloat[PCA Pursuit, sparse part, third component]{
		\includegraphics[width=\figurewidth ]{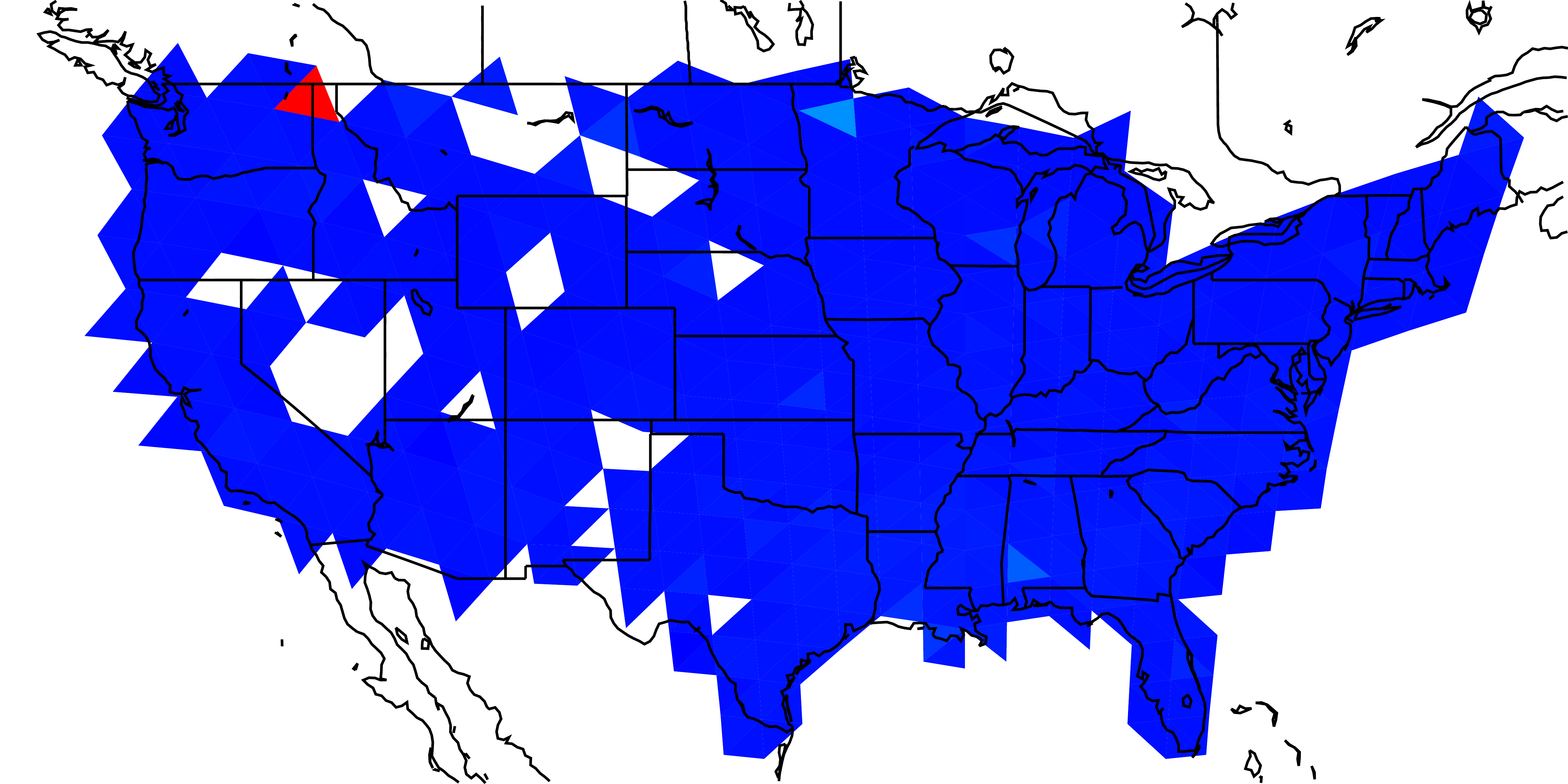}
	}
	\quad
	\subfloat[PCA Pursuit, low rank part, third component]{
		\includegraphics[width=\figurewidth ]{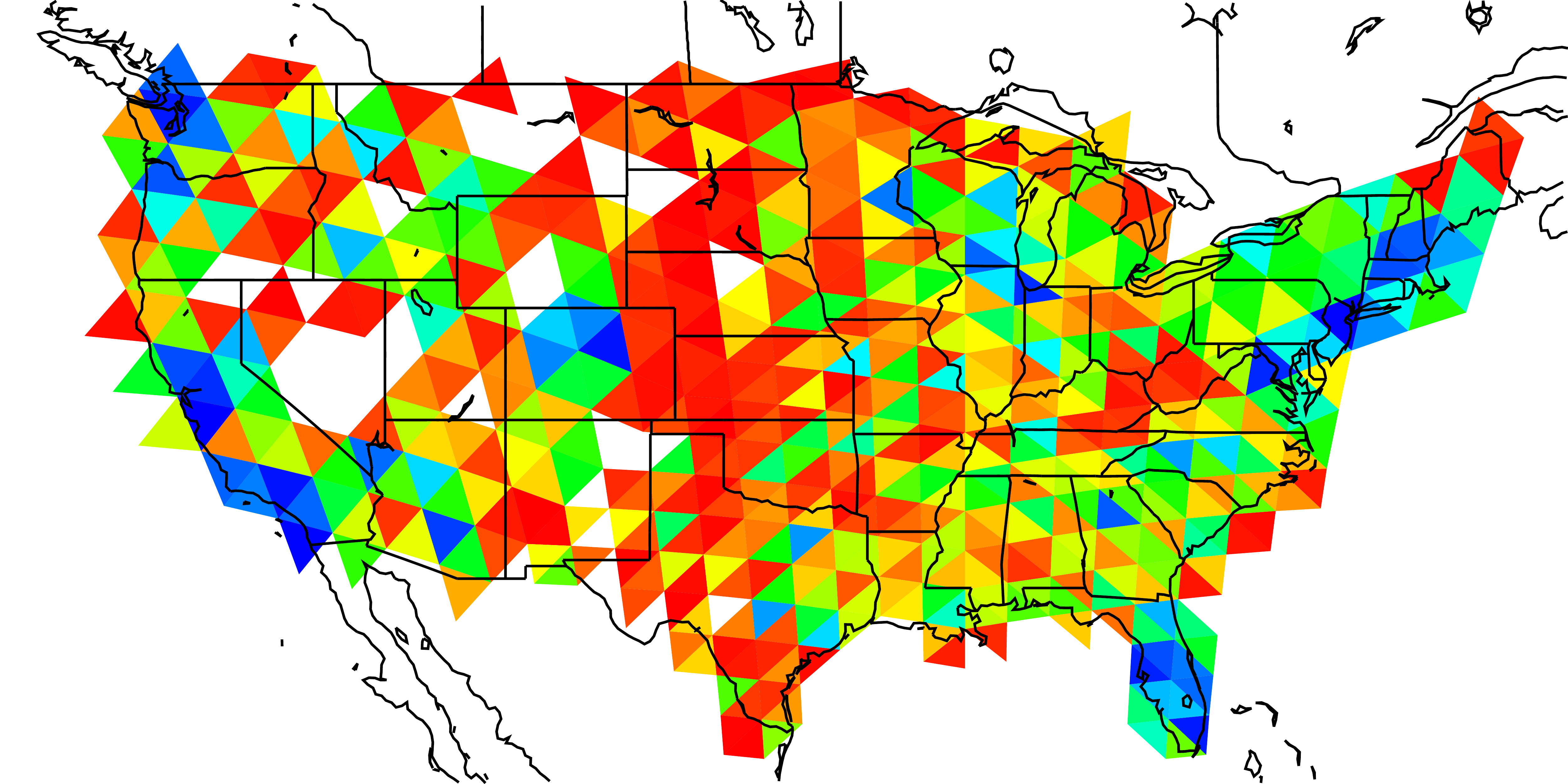}
	}
	\\
		\subfloat[Classic PCA, 4th component]{
		\includegraphics[width=\figurewidth ]{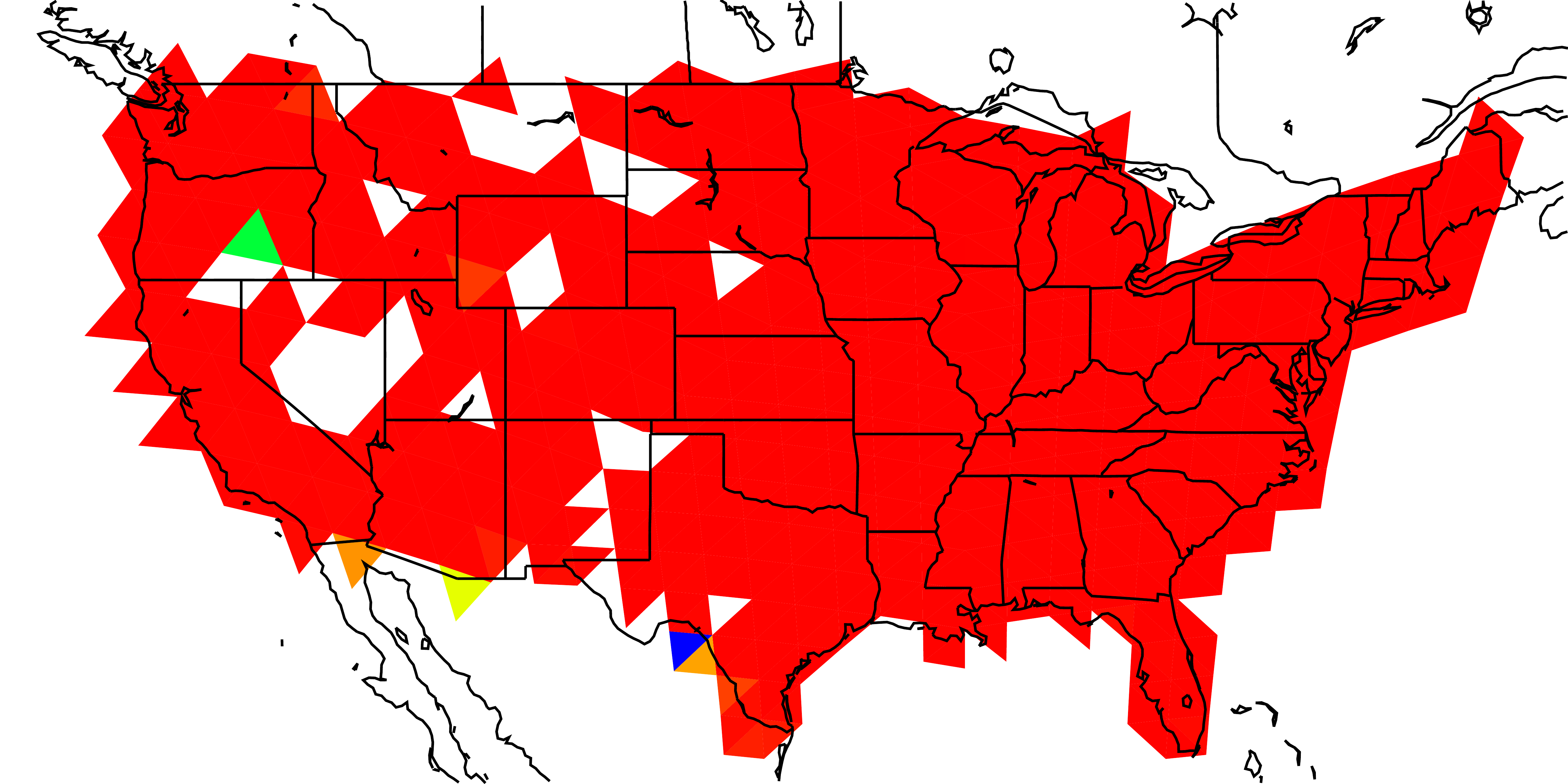}
	}
	\quad
	\subfloat[PCA Pursuit, sparse part, 4th component]{
		\includegraphics[width=\figurewidth ]{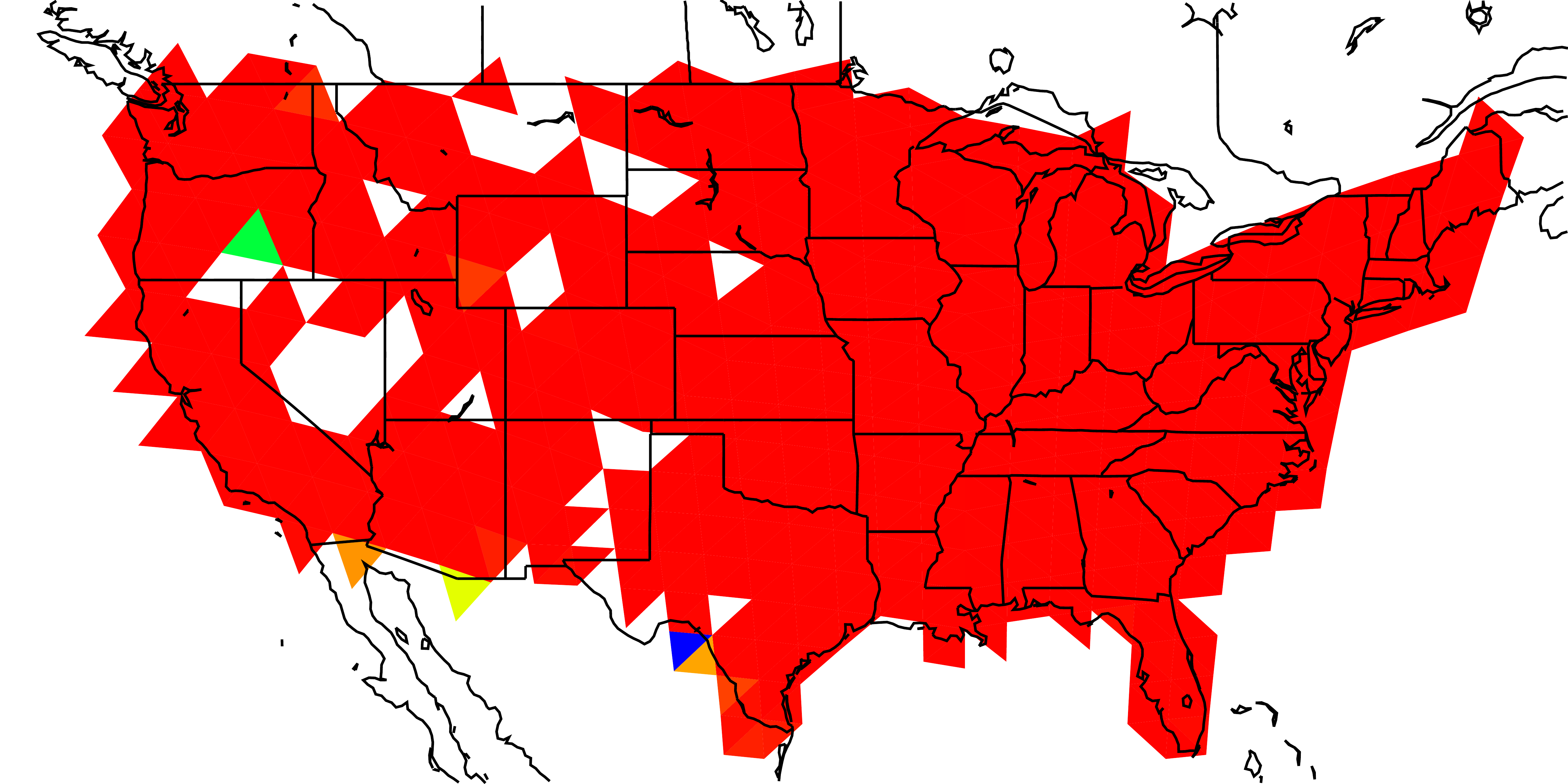}
	}
	\quad
	\subfloat[PCA Pursuit, low rank part, 4th component]{
		\includegraphics[width=\figurewidth ]{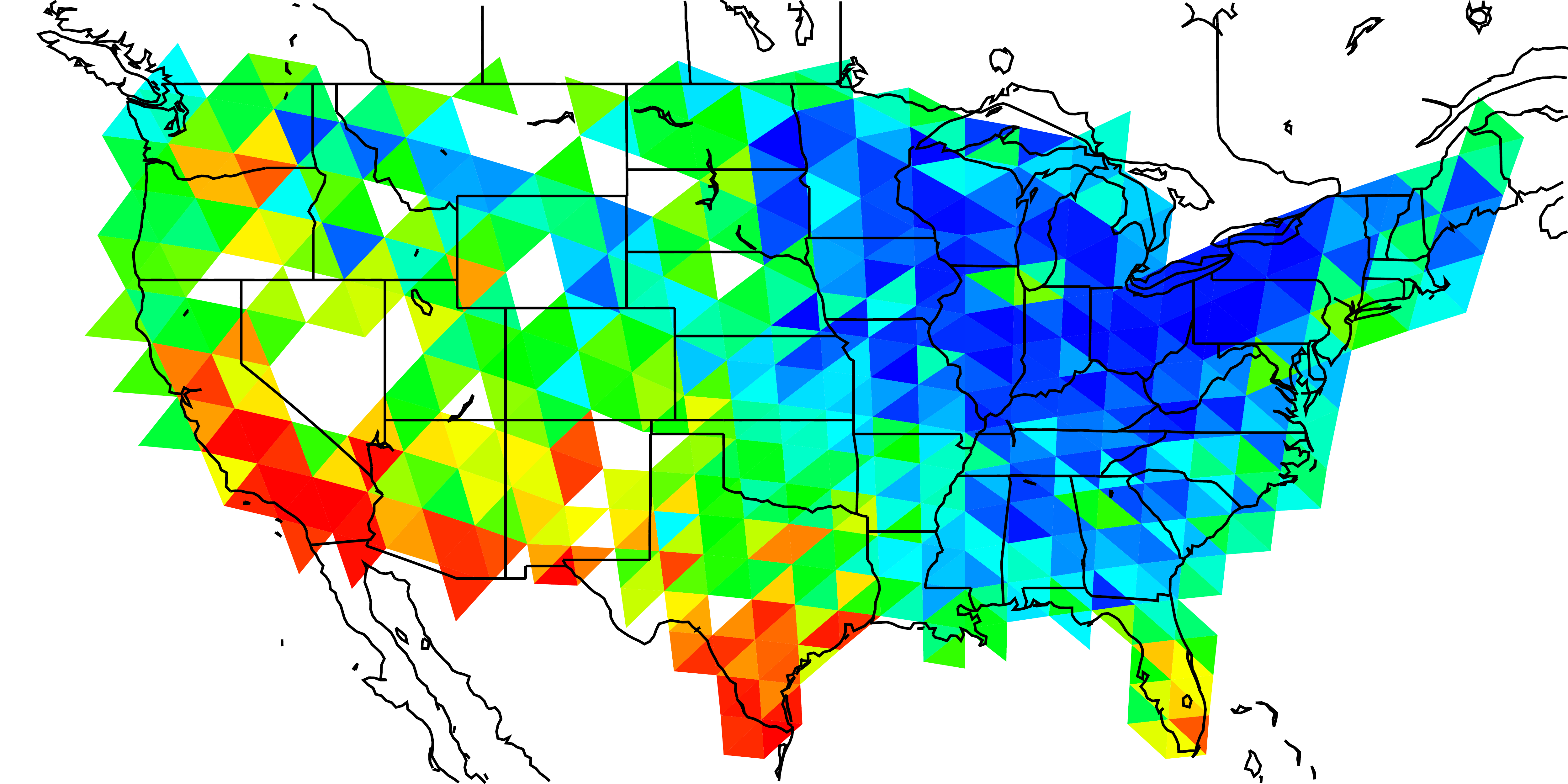}
	}
	\\
		\subfloat[twV5V5][Classic PCA, 5th component]{
		\includegraphics[width=\figurewidth ]{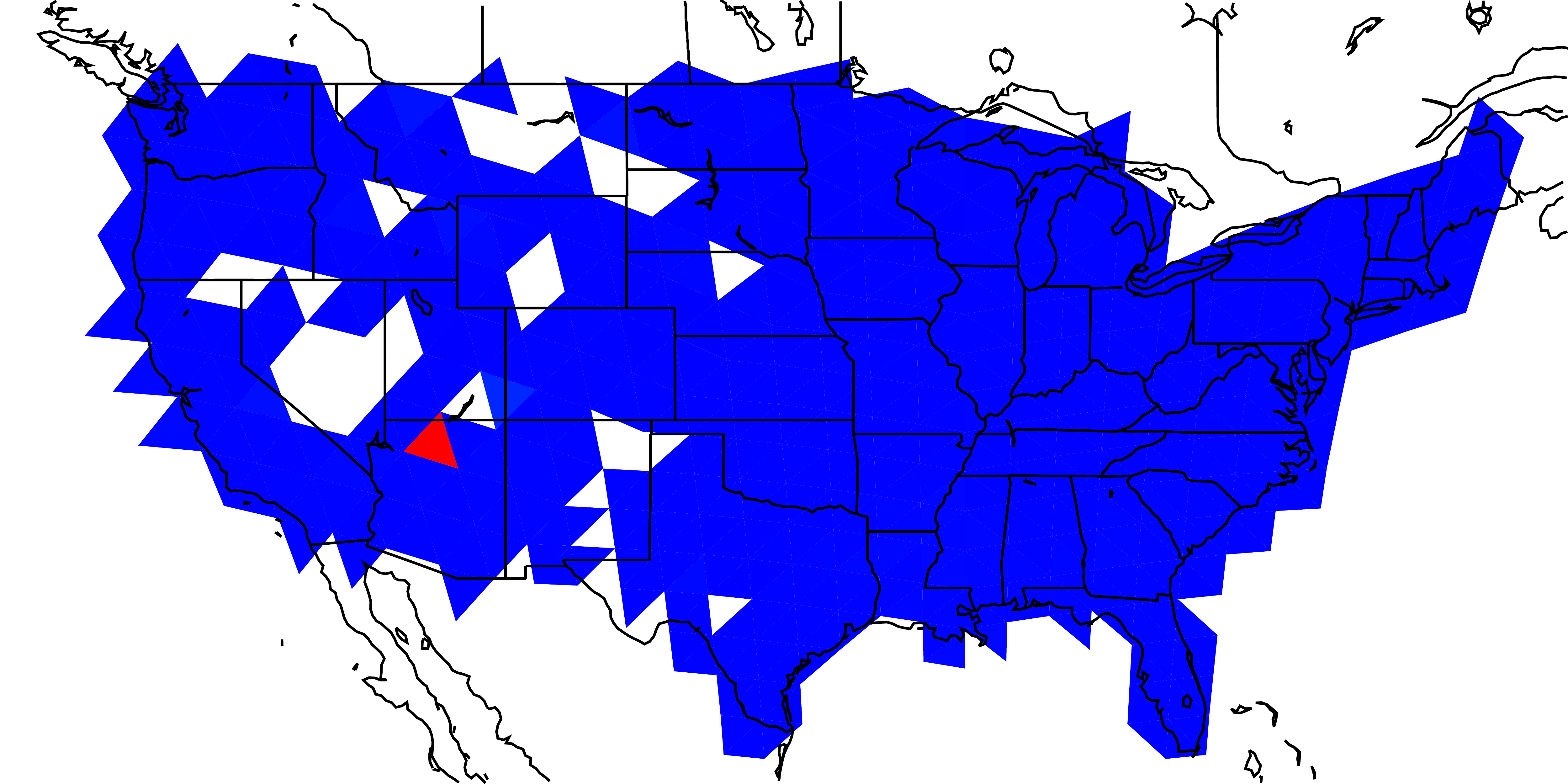}
		\label{twV5V5}
	}
	\quad
	\subfloat[PCA Pursuit, sparse part, 5th component]{
		\includegraphics[width=\figurewidth ]{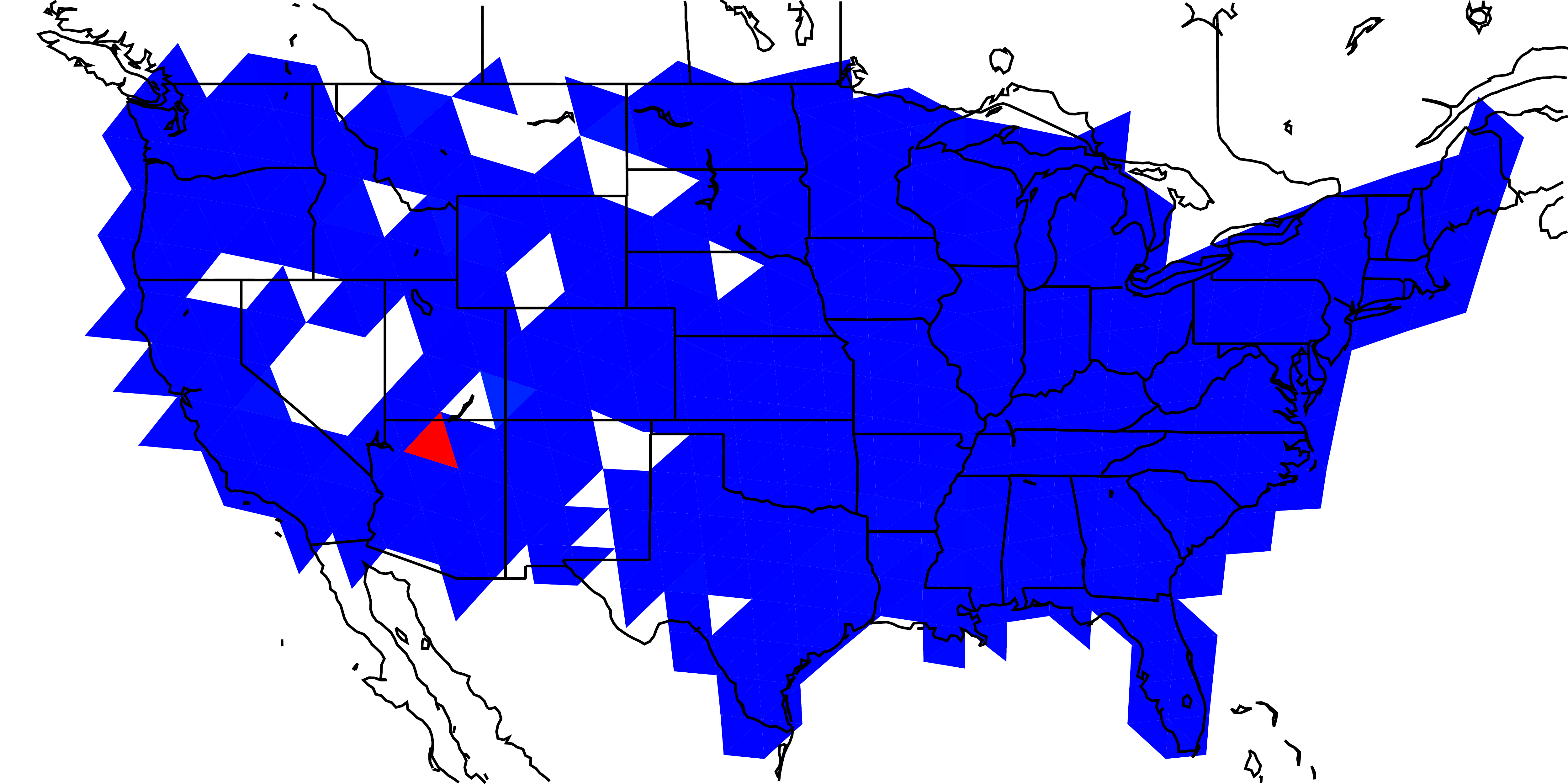}
		\label{twV5Sp5}
	}
	\quad
	\subfloat[PCA Pursuit, low rank part, 5th component]{
		\includegraphics[width=\figurewidth ]{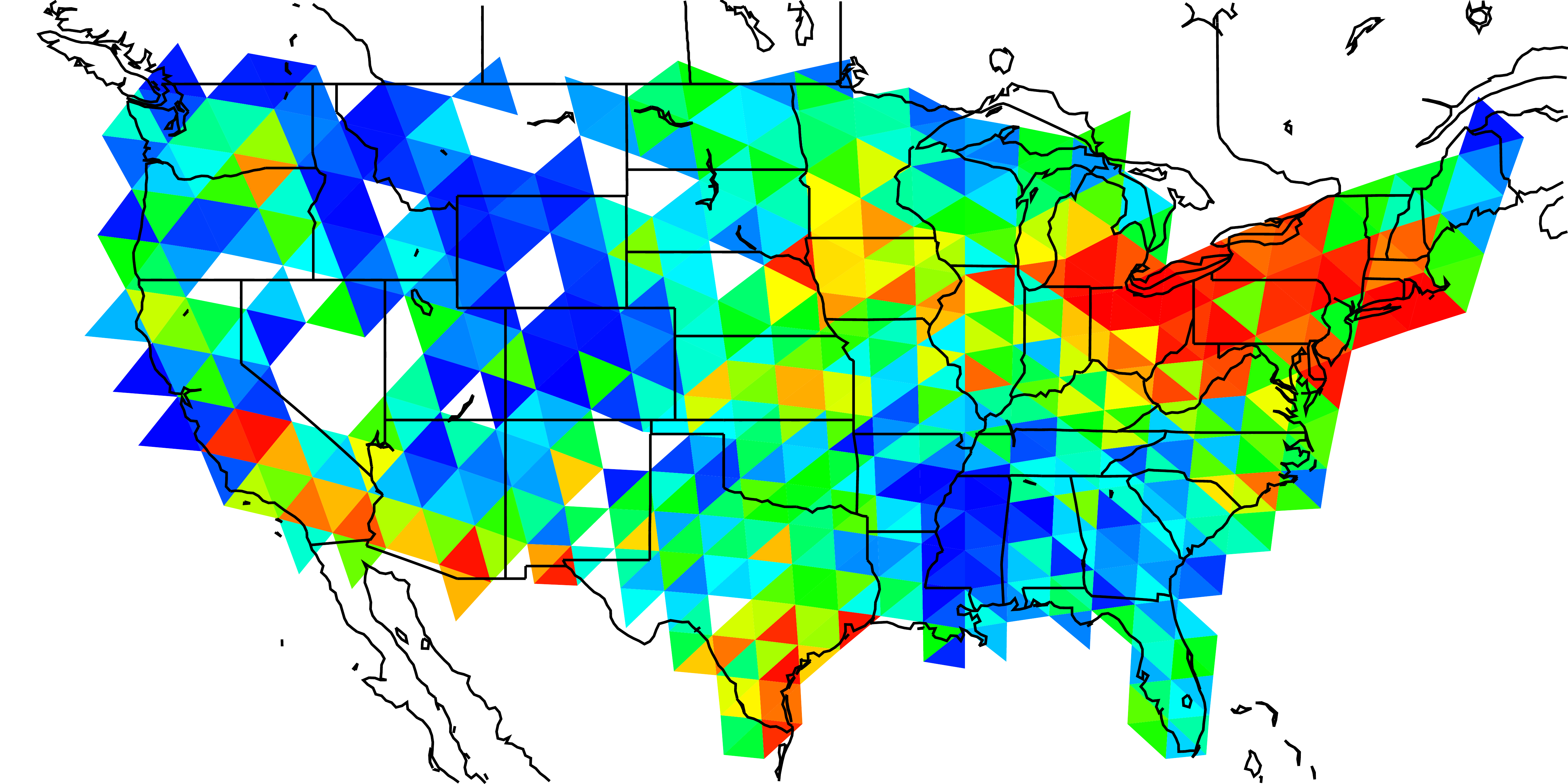}
	}
	\caption{Results of classic PCA and PCA Pursuit, first five components. The full colorscale is shown on the top. While the results of
		the classic PCA approach are mostly dominated by outlier cells, PCA Pursuit is able to separate these from the relevant
		low-dimensional geographic structure.}
	\label{twV5}
\end{figure}

\begin{table}
\begin{adjustwidth}{-2cm}{-2cm}
\centering
\scriptsize
	\subfloat[%
		Sparse part. Words giving large contributions are usually only relevant to a few cells. A significant type of these words is the ones %
		relevant to meteorological services; these arise from low-density cells where the tweets of a local meteorological service form a %
		significant proportion of all geolocated tweets. Other relevant contributions arise from tourist locations (e.g.: ``grand'' for %
		the Grand Canyon.%
	]{
	\begin{tabular}{m{1.1cm}|m{2.9cm}|m{2.9cm}|m{2.9cm}|m{2.9cm}|m{2.9cm}}
		& 1st component & 2nd component & 3rd component & 4th component & 5th component\\ \hline
		\includegraphics[width=0.7cm]{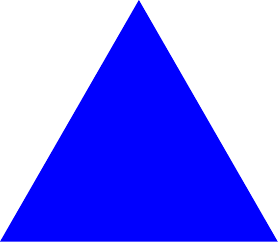} positive scores %
&	backstage, ayo, barcelona, noe, elle, encore, toronto, hosted, att, diablo, nova, gabriel, luxury, vine, ballet, diana, hbd, platform, visual, featuring
&	lol, she, ass, nigga, ain, shit, lmao, bout, tho, gone, damn, smh, dat, bitch, wit, ima, lil, wanna, niggas, gotta
&	que, for, grand, los, por, beach, con, para, day, time, del, las, park, not, national, una, best, others, night, great
&	que, los, por, con, para, del, las, una, como, pero, lol, esta, bien, todo, mas, hoy, cuando, ser, vida, hay
&	for, not, don, people, love, que, know, someone, one, night, tonight, why, wish, school, gonna, going, ever, fucking, hate, guys
\\ \hline
		negative scores \begin{center} \includegraphics[width=0.7cm]{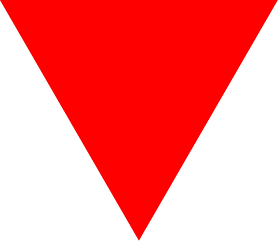} \end{center}%
&	night, think, right, need, shit, going, today, people, she, back, day, one, time, know, good, don, love, not, lol, for
&	que, underground, current, direction, humidity, new, valley, temperature, mph, great, national, winter, november, grand, wind, park, until, weather, december, for
&	gone, until, bout, today, december, she, ain, ass, nigga, lmao, rain, lol, direction, current, underground, weather, mph, temperature, humidity, wind
&	great, solutions, going, lake, right, december, state, last, bed, one, good, people, night, until, beach, not, day, tonight, today, for
&	trail, valley, december, bout, desert, village, ass, ain, lmao, nigga, view, arizona, others, posted, photo, point, lol, park, national, grand
	\end{tabular}
	}
	\\
	\subfloat[%
		Low-rank part. Here, four components represent a large-scale language usage differences throughout the USA. %
		See the text for more discussion.%
	]{
	\begin{tabular}{m{1.1cm}|m{2.9cm}|m{2.9cm}|m{2.9cm}|m{2.9cm}|m{2.9cm}}
		& 1st component & 2nd component & 3rd component & 4th component & 5th component\\ \hline
		\includegraphics[width=0.7cm]{figs/trred} positive scores %
&	disneyland, asu, utc, hazard, jacksonville, holland, rouge, boa, milwaukee, dans, graffiti, domingo, pennsylvania, dell, sur, vine, partly, personas, sterling, felicidades
&	awesome, guys, actually, idea, excited, snow, favorite, sounds, holy, doesn, amazing, totally, weird, anyone, isn, winter, huge, seriously, awkward, probably
&	grill, starbucks, mall, others, sushi, fitness, club, market, downtown, target, apartments, spa, office, lounge, pub, ale, village, arts, blvd, ave
&	paper, tanning, sweatpants, terrible, basement, pumped, awful, sitting, til, florida, cannot, study, sometime, pants, studying, classes, clearly, motivation, entire, figured
&	trip, headed, check, folks, enjoyed, enjoying, loving, wonderful, catch, app, yep, riding, service, above, missing, local, mobile, plan, choose, afternoon
			\\ \hline
		negative scores \begin{center} \includegraphics[width=0.7cm]{figs/trblue} \end{center}%
&	home, people, need, life, think, back, really, best, one, new, night, don, lol, know, for, time, good, not, day, love
&	sleepy, wen, smh, imma, tryna, females, kno, swear, goodmorning, wat, gon, dnt, oomf, avi, ima, niggas, hoes, hoe, aint, yall
&	without, either, choose, weather, smile, knew, piss, matter, laugh, find, december, won, everything, truck, saying, mine, until, quit, warning, wouldn
&	mucho, gracias, eso, mas, una, nos, como, todo, los, estoy, para, nada, pero, que, bien, por, esta, las, bueno, con
&	whos, wants, infront, alll, asshole, slut, dads, whenever, whore, asleep, hes, boyfriends, shes, youre, everyones, honestly, anymore, annoying, annoyed, texts
	\end{tabular}
	}
	\caption{Significant words identified in the results of the principal pursuit method carried out for the whole USA
		(see Fig.~\ref{twV5}).}
	\label{twU}
\end{adjustwidth}
\end{table}

On Fig.~\ref{twV5}, we show the first five principal components obtained by both methods overlayed on the map of the USA. As we deliberately
chose not to subtract any mean values from the data, the first component does not seem to contain any relevant structure. As the mean values
in the original data matrix are the same for each cell, the outlier cells displayed here are the ones where the otherwise rare words are in
abundance. The second component shows a clear structure in both cases, with the southern USA being separated from the rest (note that the
sign is inverted in the case of the low-rank part). Examining the words giving the largest contribution (i.e.~the columns of the $U$
matrix), we can identify that this separation is mainly due to swear words, abbreviations and words with spelling specific to
online social networks (see Table~\ref{twU}). In the case of the classic PCA and the sparse part identified by PCA pursuit, words
found in weather reports are also significant. These contributions mainly come from the northern USA, where in some cells a large proportion
of geolocated tweets come from meteorological services. On the other hand, in the case of the low-rank part identified by the PCA pursuit,
the southern USA is mainly counteracted by words relevant to the geographic features of the northern USA, e.g. ``winter'' or ``snow''.
This means that the separation is effective, the ``outlier'' cells are indeed well filtered by the method.

In the case of the higher components, the difference between the two methods becomes more drastic. The results of the classic PCA are
dominated by outliers -- cells which are isolated centers where the content of the tweets differ significantly from the rest. Some of
these can be considered ``noise'': meteorological stations, service announcements and tweets of job advertisements; on the other hand,
some of these show real localized features: e.g. in the case of fifth component, the ``outlier'' cell is the location of the Grand Canyon,
where a significant proportion of tweets indeed mention the canyon itself (see Figs.~\ref{twV5V5} and~\ref{twV5Sp5}). In the case of the
PCA Pursuit, these results are present in the sparse part (which is very similar to the results of the classic PCA). The low-rank part
contains slowly varying vectors, which mimic of the geographic and social features of the USA.

In the case of the low-rank part, in the third component, the major cities are separated from the countryside: words like 'downtown', 'sushi',
'mall' or 'pub' are opposed by words like 'truck'. In the case of the fourth component, areas with high ratio of Spanish speaking Twitter
users are separated from the Northeastern USA.

\subsection{New York City}

\begin{figure}
	\centering
	\subfloat[1st component]{
		\includegraphics[width=\figurewidthNY ]{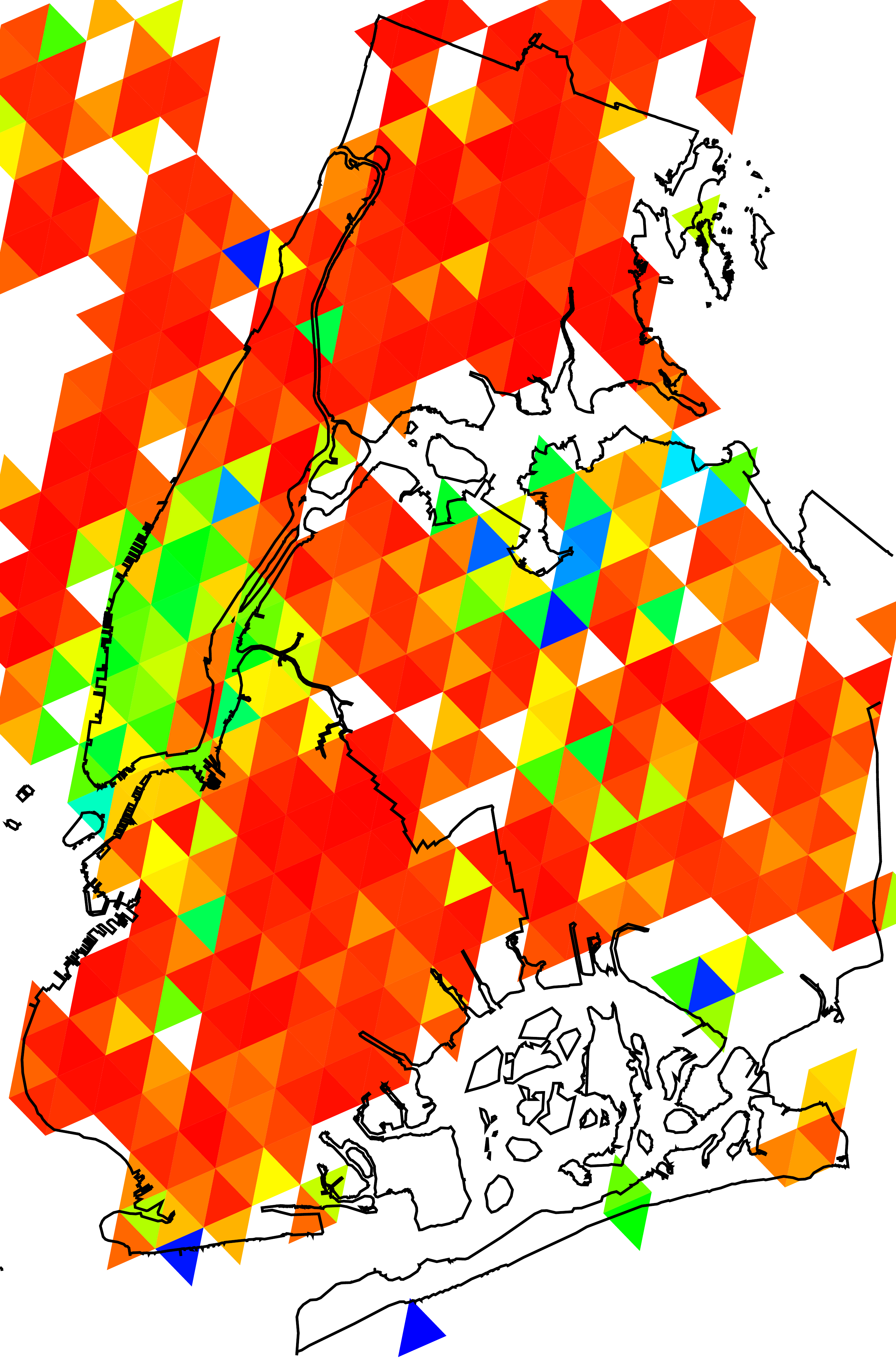}
	}
	\subfloat[2nd component]{
		\includegraphics[width=\figurewidthNY ]{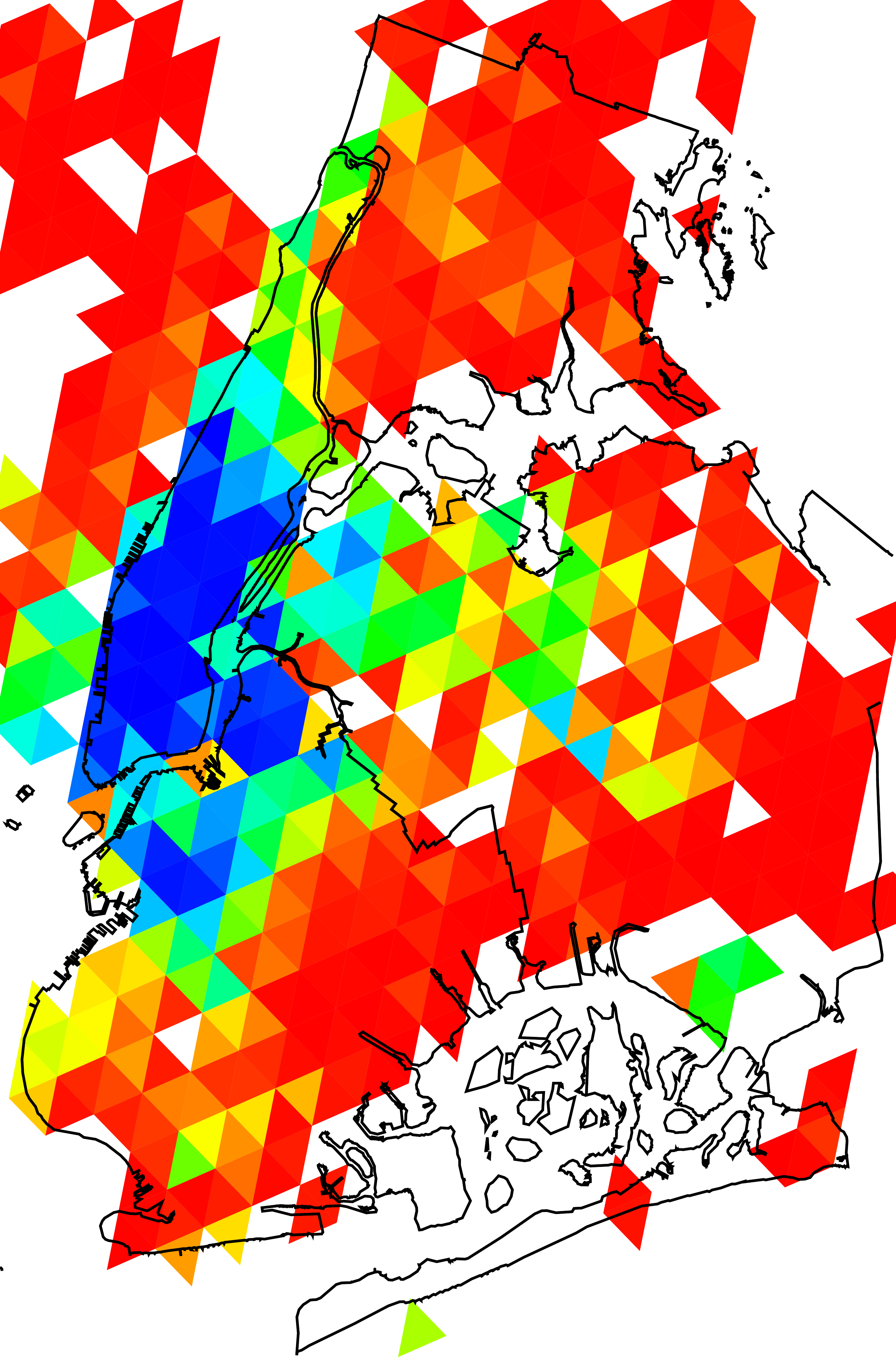}
	}
	\subfloat[3rd component]{
		\includegraphics[width=\figurewidthNY ]{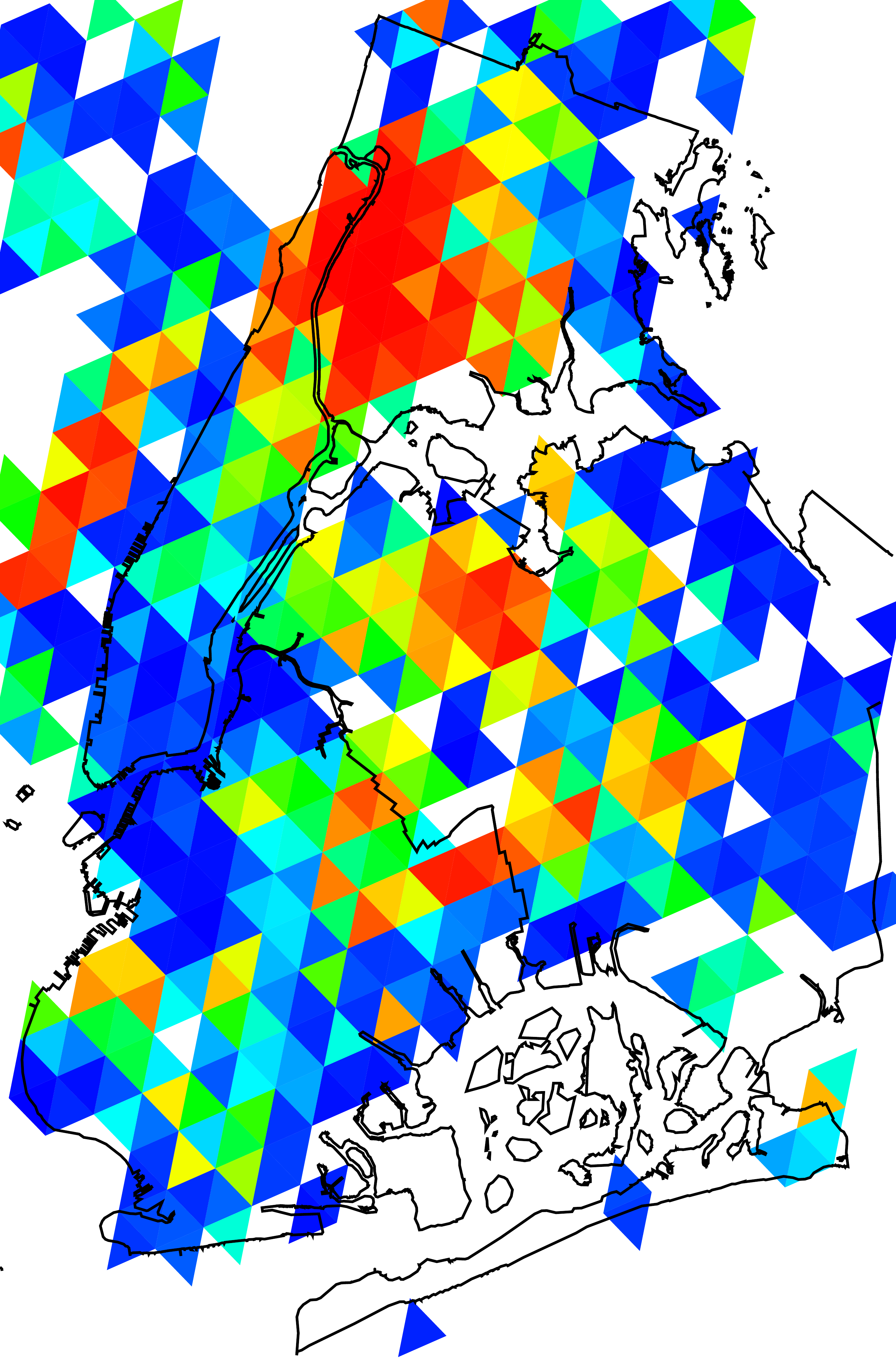}
	}
	\subfloat[4th component]{
		\includegraphics[width=\figurewidthNY ]{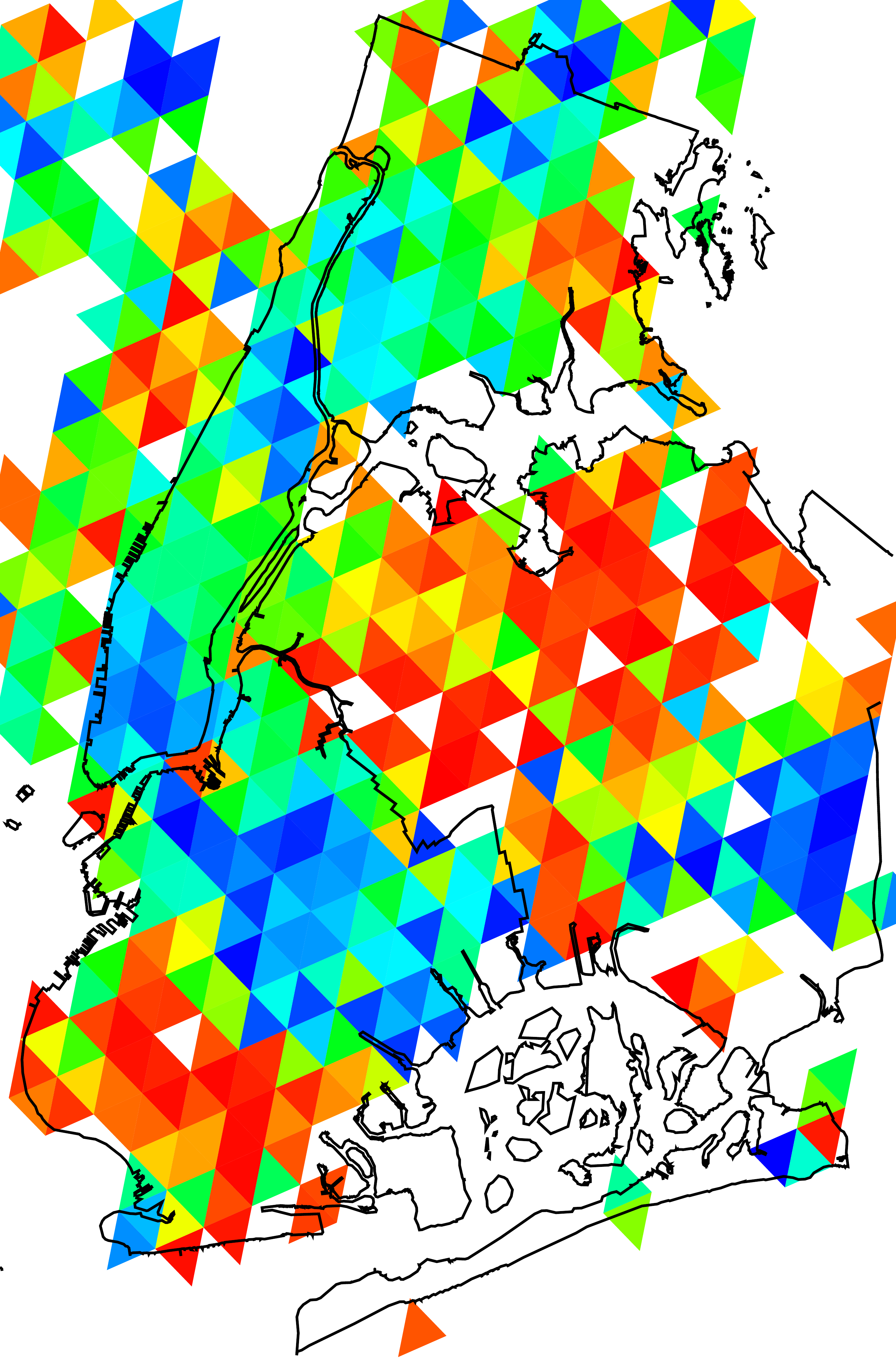}
	}
	\subfloat[5th component]{
		\includegraphics[width=\figurewidthNY ]{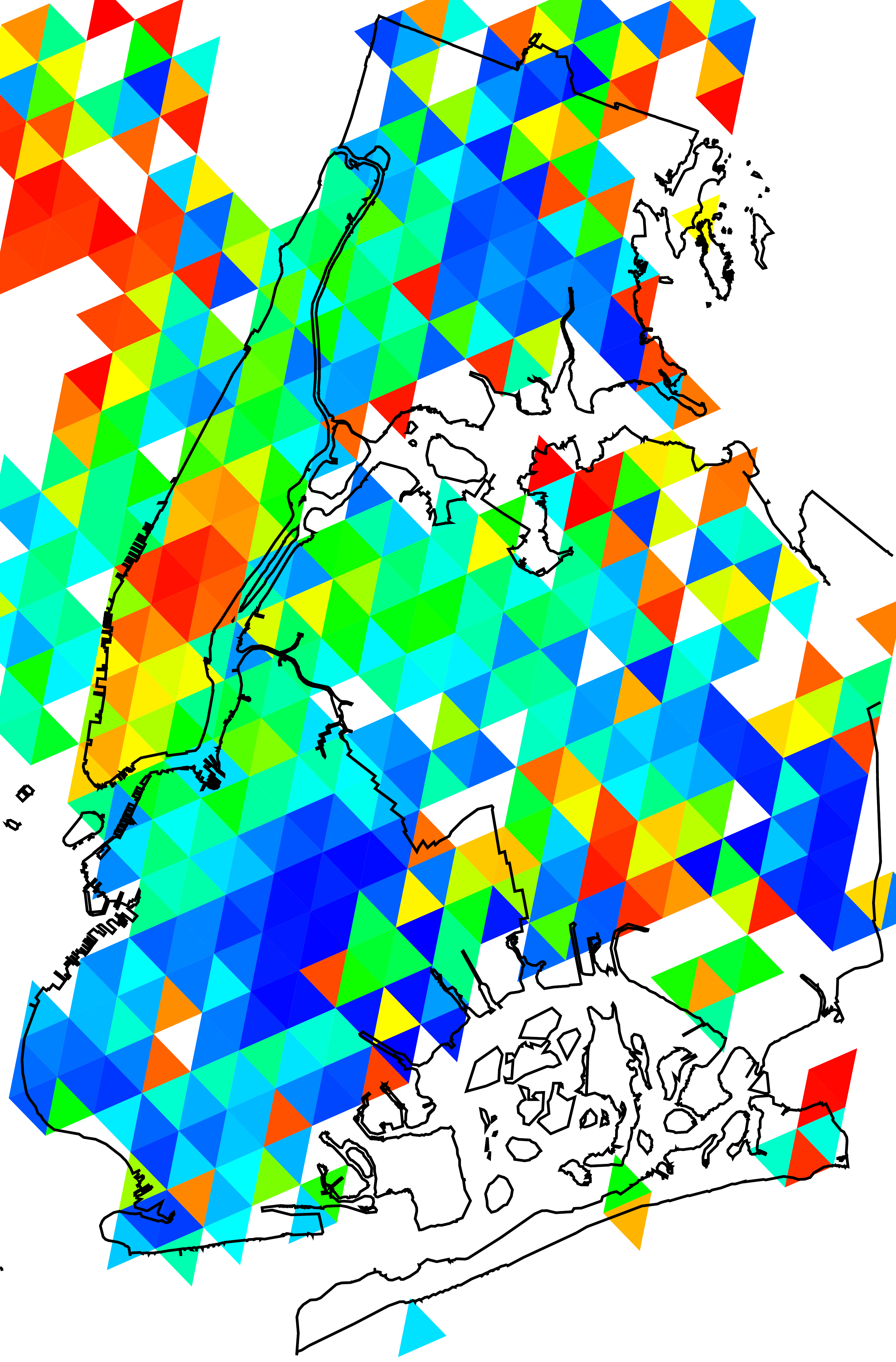}
	}
	\caption{Results of PCA Pursuit applied to New York city, first five components, low-rank part. The results roughly partition the
		city to its boroughs.}
	\label{nyV5}
\end{figure}

\begin{table}
\begin{adjustwidth}{-2cm}{-2cm}
\scriptsize
\centering
	\begin{tabular}{m{0.8cm}|m{3cm}|m{3cm}|m{3cm}|m{3cm}|m{3cm}}
		& 1st component & 2nd component & 3rd component & 4th component & 5th component\\ \hline
		\includegraphics[width=0.7cm]{figs/trred} positive scores %
&	armory, guggenheim, bergen, flatiron, pianos, saks, consulting, jetblue, seaport, gramercy, intrepid, pinkberry, caffe, herald, richmond, cinemas, uniqlo, saloon, katz, championships
&	coffee, manhattan, awesome, brunch, others, bar, photo, nyc, apartment, street, subway, posted, totally, cafe, ousted, film, beer, also, view, 2012
&	actually, power, apparently, two, also, hour, found, first, place, year, seriously, three, literally, kind, sure, holy, least, dinner, week, run
&	development, gallery, cocktails, sidewalk, une, ipa, intern, espresso, panel, vegan, screening, venue, pas, avec, cocktail, companies, filming, ceo, designer, dans
&	records, corn, amy, trains, sake, beyonce, oven, recommend, image, potato, ultimate, charles, impressed, rolls, trailer, artists, madonna, cap, homemade, roommate
\\ \hline
		negative scores \begin{center} \includegraphics[width=0.7cm]{figs/trblue} \end{center}%
&	going, today, right, lol, see, will, got, don, love, good, know, just, like, get, one, now, back, new, time, day
&	funny, cant, ima, smh, yea, lmaoo, hate, niggas, gotta, sleep, dont, mad, text, tho, cause, ass, wanna, bitch, nigga, idk
&	voy, muy, tienes, mejor, para, nada, esa, estoy, siempre, tiene, cuando, una, mas, por, como, las, pero, eso, hoy, esta
&	shore, yay, picking, island, hahaha, sucks, luck, excited, coffee, hours, hour, holy, mets, car, guys, rangers, awesome, dad, amazing, haha
&	bistro, institute, het, offices, nao, oyster, reade, lga, meetup, sono, een, gotham, tribeca, airlines, falafel, meu, minha, stout, vai, vou
	\end{tabular}
	\caption{Important words identified in the low-rank part of the PCA Pursuit for New York City.}
	\label{nyU}
\end{adjustwidth}
\end{table}

\begin{figure}
	\centering
	\subfloat[1st component]{
		\includegraphics[width=\figurewidthNY ]{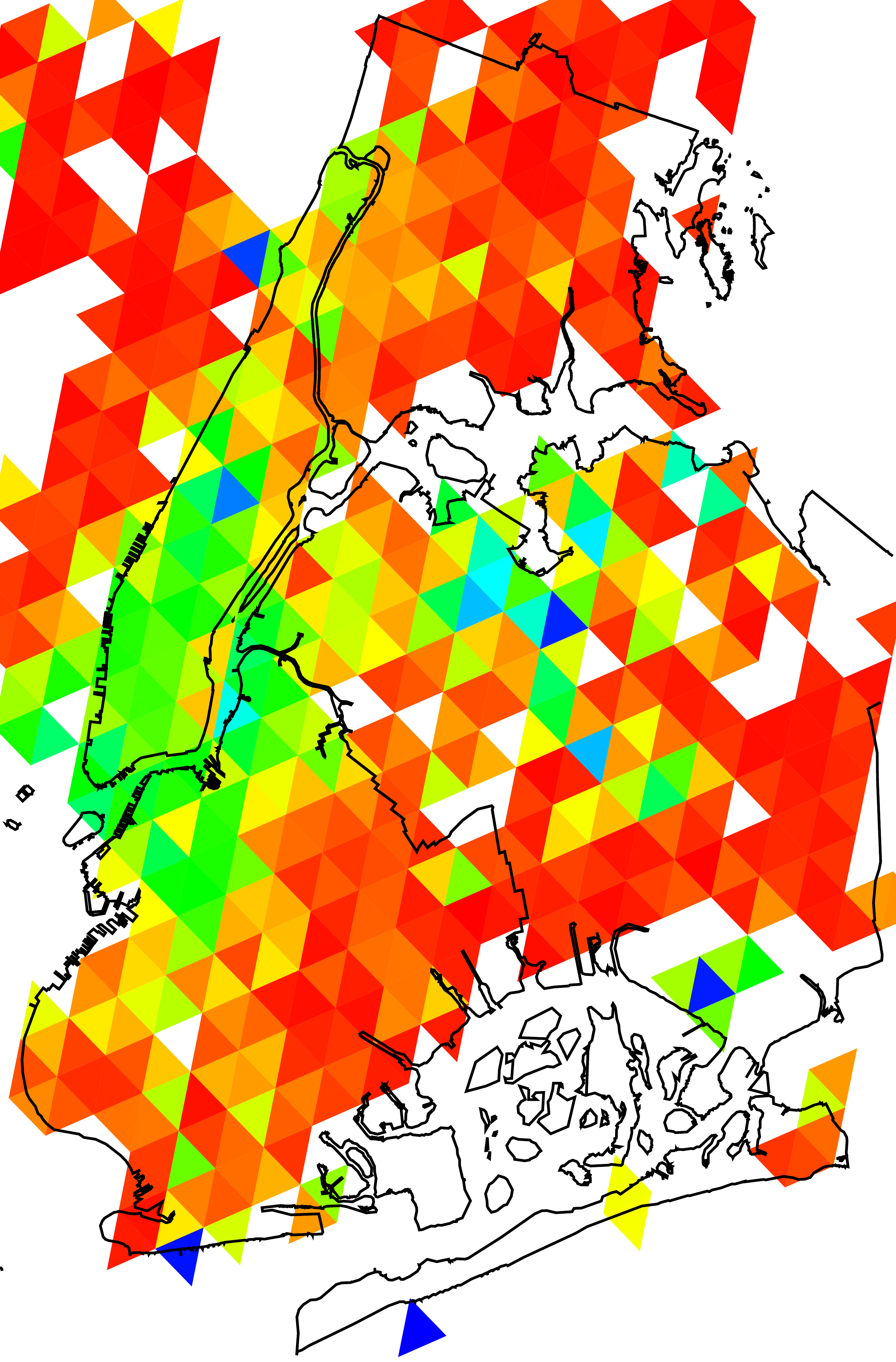}
	}
	\subfloat[2nd component]{
		\includegraphics[width=\figurewidthNY ]{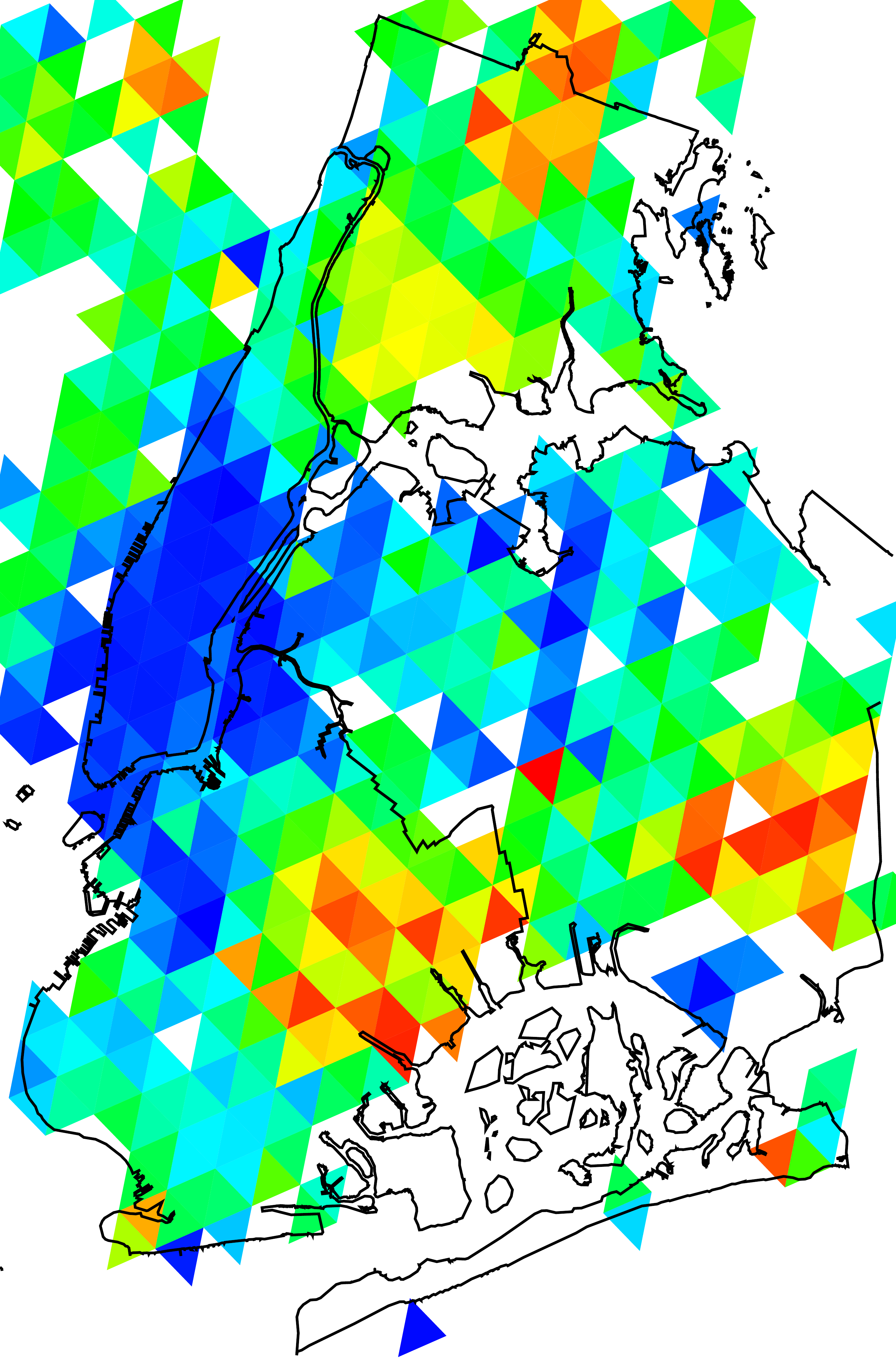}
	}
	\subfloat[3rd component]{
		\includegraphics[width=\figurewidthNY ]{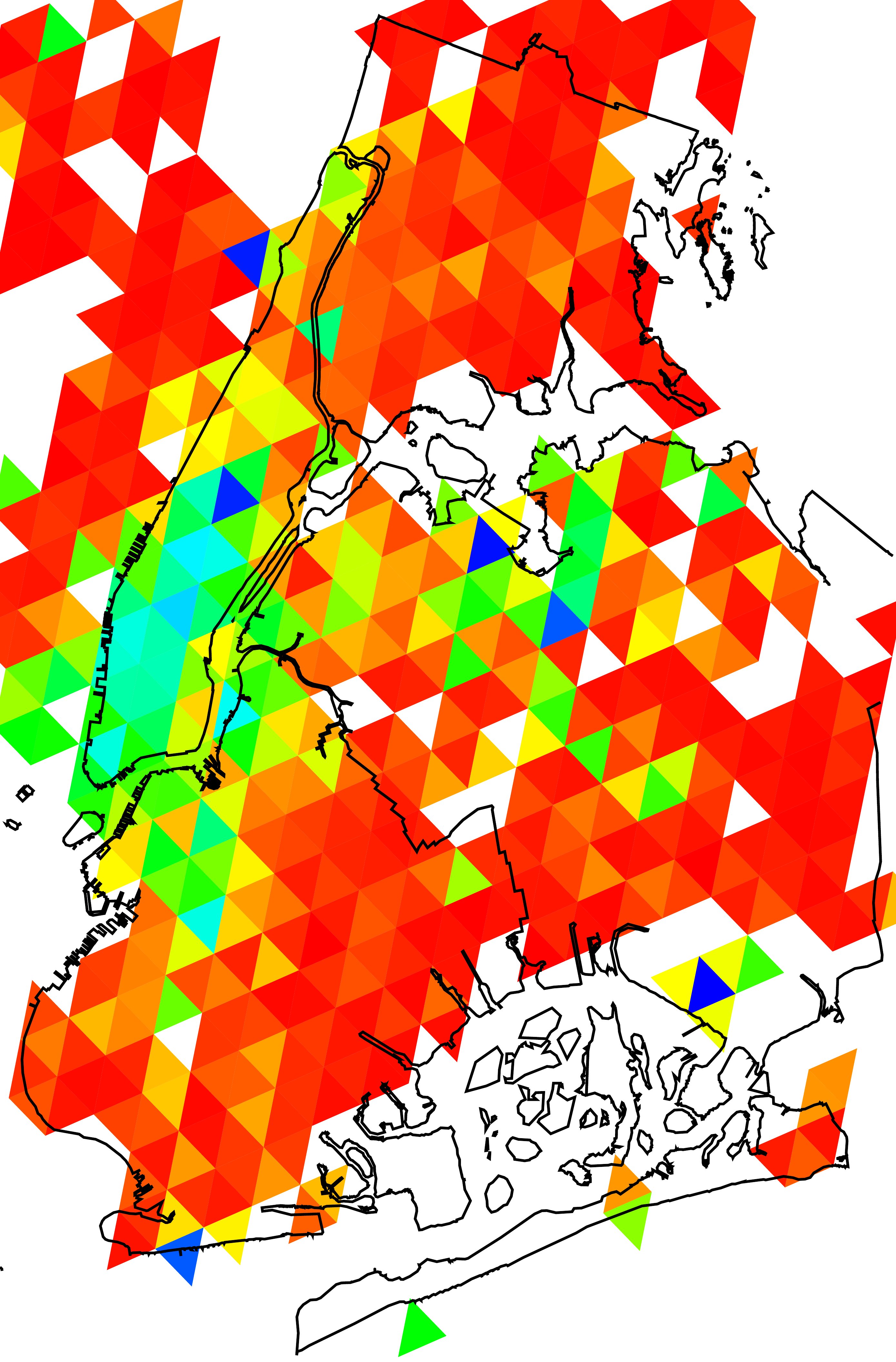}
	}
	\subfloat[4th component]{
		\includegraphics[width=\figurewidthNY ]{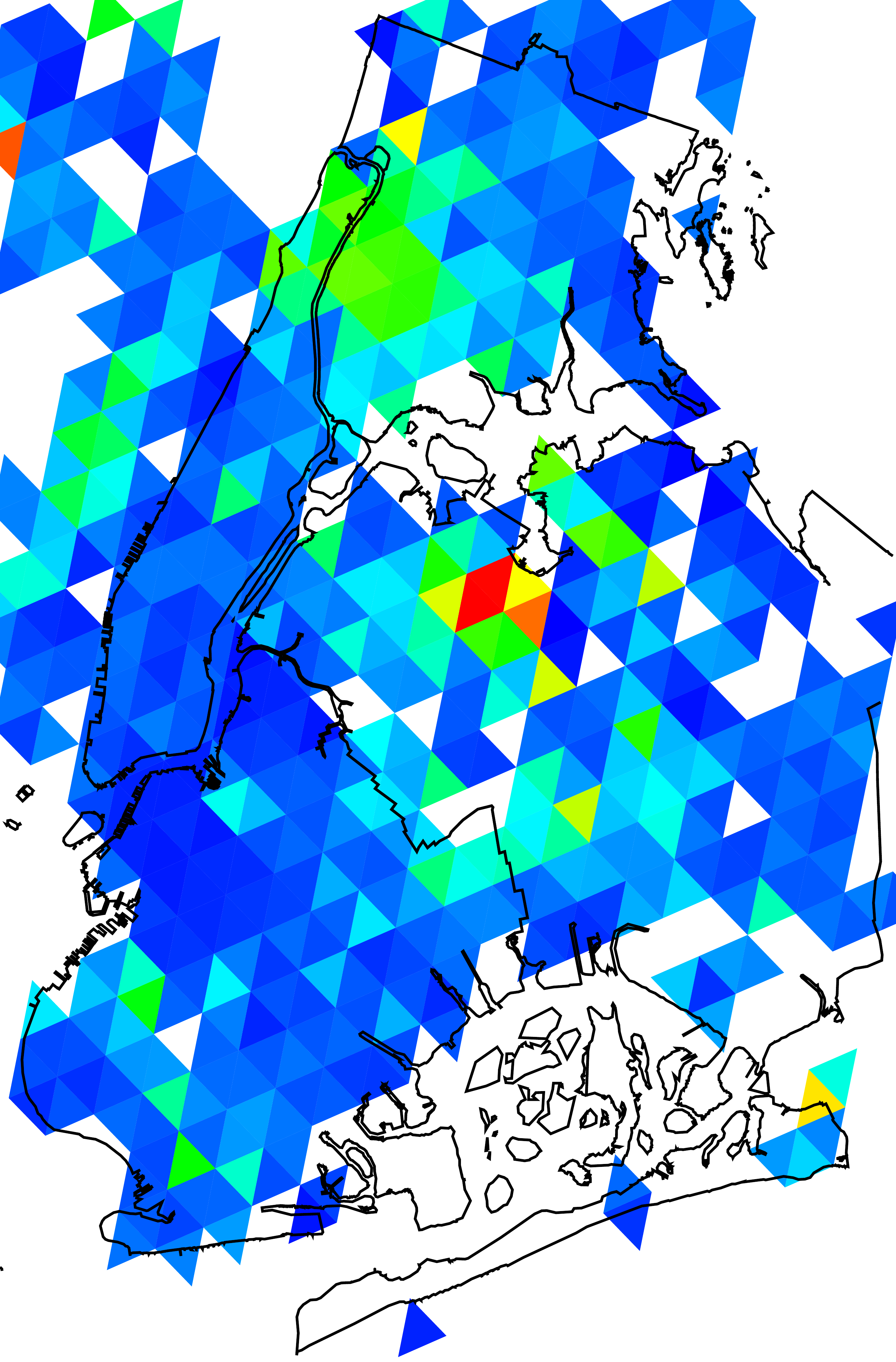}
	}
	\subfloat[5th component]{
		\includegraphics[width=\figurewidthNY ]{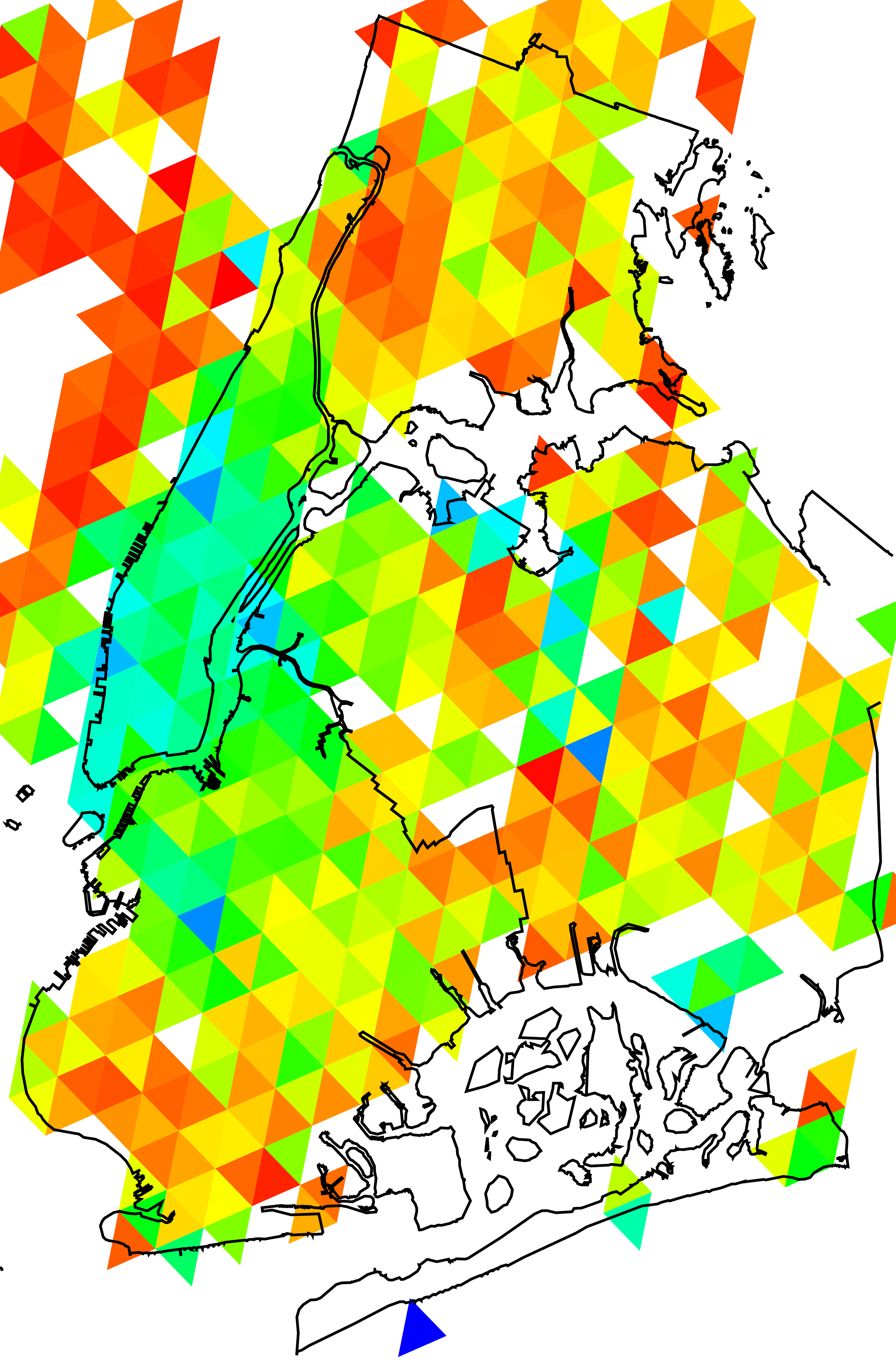}
	}
	\caption{The tweets of New York City, projected to the space defined by the first five principal components obtained from applying
		PCA Pursuit to the whole USA (low-rank part).}
	\label{nytwV5}
\end{figure}

We have seen that we were able to identify the low-rank structure of variations in language use on the scale of the USA. To test
if there are such variations present on a much smaller scale, we applied the same method to the region containing most of New York
City (Fig.~\ref{nycntl}). Results obtained for this corpus are depicted on Fig.~\ref{nyV5} (principal components of the obtained
low-rank matrix). These components separate the city into its various districts. In the case of the first two components, Manhattan
is mainly separated from the rest of the city; in the case of the second component, the separation is again in part due to the use
of swear words and online text specific abbreviations. The third component is now mainly about the ratio of Spanish texts, with the
Bronx and Queens giving significant contributions.

We also examine whether the space of principal components obtained for New York City is similar to the components obtained
for the whole USA. We project the word occurrence matrix of New York City into the space of principal components obtained for
the whole USA: $V^{\textrm{(proj)}} = U^{\textrm{(USA)}} X^{\textrm{(NY)}}$, and then plot the obtained vectors on the map
of New York (Fig.~\ref{nytwV5}). We get that most components separate the city in a meaningful way, suggesting that some of
the differences obtained at large scales are relevant on much smaller scales too.

\section{Conclusions and future work}

In this paper, we investigated the feasibility of principal component analysis to identify regional features of user content
present in geotagged Twitter messages. Using the \emph{PCA Pursuit} technique, we were able to separate the low-rank and sparse
parts present in the data, and identify some of the main features in both. Examining the spatial distribution of the principal
components of the low-rank part, we found that there are indeed large-scale spatial variations present among Twitter users.
We also investigated the scalability of this method and found that some of the principal components found for the whole USA
are also relevant if we only consider the much smaller area of New York City.

The methods presented here can be the basis for studying language use on large-scale in the future. The ability to separate
the word usage matrix into a low-rank and sparse component opens up many possibilities. The principal components of the sparse
part can be used to identify regional or highly localized topics of interest, and also to identify sources of Twitter content
generated by bots and services, as these can be easily spotted in otherwise low-density regions. On the other hand, the low-rank
part shows relevant regional variations in language use. Exploring possible connections between language use differences revealed
by the methods employed here and other demographic or social features will probably yield new insights into the dynamics of
society and language. Our next goal is to combine the PCA Pursuit method with sentiment analysis methods developed previously by
Mitchell~et~al.~\cite{hedonometrics}.

On the other hand, the results of PCA might lead to more direct applications. Identifying locations of users based on the content
of their messages has been previously evaluated using probabilistic approaches~\cite{localization1}. Now, using the space of
PCA components, we might be able to estimate the location of texts (e.g.~a small set of tweets from a specific user, or
connected group of users) more efficiently. Also, comparing the result obtained here with the results obtained by text mining
methods focusing on other (i.e.~non geographic) features has the potential to extend those approaches to obtain spatial
results too.

Also, our further goals include implementing the better integration of the PCA Pursuit method and the related data processing steps into the
database to obtain an integrated and easy-to-use environment for analyzing large corpuses of possibly geo-tagged texts.

\section*{Acknowledgments}

	The authors thank the partial support of
	the European Union and the European Social Fund through project FuturICT.hu
	(grant no.: TAMOP-4.2.2.C-11/1/KONV-2012-0013), the OTKA 7779 and the NAP 2005/KCKHA005 grants.
	EITKIC\_12-1-2012-0001 project was partially supported by the Hungarian Government,
	managed by the National Development Agency, and financed by the Research and
	Technology Innovation Fund and the MAKOG Foundation.


\begin{thebibliography}{99}

	\bibitem{lsalandauer}
		Landauer, T., and Dumais, S. (1997).
		A solution to Plato’s problem: The latent semantic analysis theory of acquisition, induction, and representation of knowledge.
		{\em Psychological review}, {\bf 1}(2), 211--240.
	
	\bibitem{semantic}
		Zsigmondi, Zs., and Kiss, A. (2013).
		Implementation of Natural Language Semantic Wildcards using Prolog.
		{\em SPLST 13} 
	
	\bibitem{lsaindexing}
		Deerwester, S., Dumais, S. T., Furnas, G. W., Landauer, T. K. and Harshman, R. (1990)
		Indexing by Latent Semantic Analysis.
		{\em Journal of the American Society for Information Science}, {\bf 41}, 391.
		
	\bibitem{lsa1}
		Foltz, P. W., Kintsch, W., and Landauer, T. K. (1998).
		The measurement of textual coherence with latent semantic analysis.
		{\em Discourse Processes}, {\bf 25}, {285--307}.
	
	\bibitem{lsa2}
		Bellegarda, J. R., Butzberger, J. W., Yen-Lu Chow, Coccaro, N. B., and Naik, D. (1996)
		A novel word clustering algorithm based on latent semantic analysis.
		{\em ICASSP-96}. 

		
	\bibitem{pcapursuit1}
		Lin, Z., Chen, M., and Ma, Y. (2010).
		The augmented Lagrange multiplier method for exact recovery of corrupted low-rank matrices.
		\texttt{arXiv:1009.5055}. 
	
	\bibitem{pcapursuit2}
		Emmanuel J. Cand\`es, Xiaodong Li, Yi Ma, John Wright (2011).
		Robust Principal Component Analysis.
		{\em JACM}, {\bf 58} (3), 11. 
		
	\bibitem{alm}
		We use the Matlab code made available by Chen~et~al.: \verb=http://perception.csl.illinois.edu/matrix-rank/sample_code.html=.
		
	\bibitem{findmefacebook}
		Backstrom, L., Sun, E., and Marlow, C. (2010).
		Find me if you can: improving geographical prediction with social and spatial proximity.
		{\em WWW  ’10: Proceedings of the 19th international conference on World wide web.} 
		
	\bibitem{localization1}
		Cheng, Z., Caverlee, J., and Lee, K. (2010).
		You are where you tweet: a content-based approach to geo-locating twitter users.
		{\em Proc. CIKM10} 
	
	\bibitem{diffusion1}
		Bakshy, E., Rosenn, I., Marlow, C., and Adamic, L. (2012).
		The role of social networks in information diffusion.
		{\em WWW '12 Proceedings of the 21st international conference on World Wide Web},
		519--528. 
	
	\bibitem{influence1}	
		Bakshy, E., and Hofman, J. (2011).
		Everyone’s an influencer: quantifying influence on twitter.
		{\em Proceedings of the fourth ACM international conference on Web search and data mining.} 
	
	\bibitem{memetracking}
		Leskovec, J., Backstrom, L., and Kleinberg, J. (2009).
		Meme-tracking and the dynamics of the news cycle.
		{\em Proceedings of the 15th ACM SIGKDD international conference on Knowledge discovery and data mining - KDD  ’09}, 497. 
	
	\bibitem{diffusion2}
		Eisenstein, J., O’Connor, B., Smith, N., and Xing, E. (2012).
		Mapping the geographical diffusion of new words.
		\texttt{arXiv: 1210.5268}
	
	\bibitem{leskovec}
		Rodriguez, M. G., Leskovec, J., and Sch\"olkopf, B. (2012).
		Structure and Dynamics of Information Pathways in Online Media.
		\texttt{arXiv: 1212.1464}
	
	\bibitem{postagging}	
		Foster, J., \c{C}etinoglu, \"O., Wagner, J., Le Roux, J., Hogan, S., Nivre, J., Hogan, D. and Van Genabith, J. (2011).
		\# hardtoparse: POS Tagging and Parsing the Twitterverse.
		{\em 25. AAAI Workshop, Analyzing Microtext.}
	
	\bibitem{influence2}	
		Myers, S., Zhu, C., and Leskovec, J. (2012).
		Information diffusion and external influence in networks.
		{\em Proceedings of the 18th ACM SIGKDD international conference on Knowledge discovery and data mining.} 
		
	\bibitem{watts}
		Watts, D., and Dodds, P. (2007).
		Influentials, networks, and public opinion formation.
		{\em Journal of consumer research}, {\bf 34}. Retrieved from http://www.jstor.org/stable/10.1086/518527

	
	\bibitem{twitterdb}
		Dobos, L., et.al. (2013).
		A Multi-terabyte Database for Geotagged Social Network Data.
		{\em Submitted to the CogInfoCom 2013}.

	\bibitem{hedonometrics}
		Mitchell L, Frank MR, Harris KD, Dodds PS, Danforth CM (2013).
		The Geography of Happiness: Connecting Twitter Sentiment and Expression, Demographics, and Objective Characteristics of Place.
		{\em PLoS ONE} {\bf 8}(5): e64417.
		
	\bibitem{htm} A. Szalay, J. Gray, Gy. Fekete, P. Kunszt, P. Kukol, and A. Thakar,
		Indexing the Sphere with the Hierarchical Triangular Mesh.
		{\em Microsoft Research Technical Report, MSR-TR-2005-123} (2005).
	

\end{thebibliography}
\end{document}